\documentclass[10pt,twocolumn,letterpaper]{article}

\usepackage[pagenumbers]{cvpr}      

\usepackage{graphicx}
\usepackage{amsmath}
\usepackage{amssymb}
\usepackage{booktabs}
\usepackage{tabularx}
\usepackage{soul}
\usepackage{float}
\usepackage{multirow}
\usepackage[multiple]{footmisc}
\usepackage{comment}
\makeatletter
\renewcommand{\paragraph}{%
  \@startsection{paragraph}{4}%
  {\z@}{1ex \@plus 0.5ex \@minus .2ex}{-1em}%
  {\normalfont\normalsize\bfseries}%
}
\makeatother

\newcolumntype{L}{>{\centering\arraybackslash}m{1.2cm}}

\usepackage[hyperfootnotes=true,pagebackref,breaklinks,colorlinks]{hyperref}

\usepackage[capitalize]{cleveref}
\crefname{section}{Sec.}{Secs.}
\Crefname{section}{Section}{Sections}
\Crefname{table}{Table}{Tables}
\crefname{table}{Tab.}{Tabs.}

\def\cvprPaperID{1482} 
\def\confName{CVPR}
\def\confYear{2022}

\begin{document}

\title{CRAFT: Cross-Attentional Flow Transformer for Robust Optical Flow}

\author{Xiuchao Sui$^1$\thanks{Equal .} \hspace{0.8em}  Shaohua Li$^1$\footnotemark[1] \hspace{0.8em} Xue Geng$^2$ \hspace{0.8em} Yan Wu$^2$ \hspace{0.8em} \\
Xinxing Xu$^1$ \hspace{0.8em} Yong Liu$^1$ \hspace{0.8em} Rick Goh$^1$ \hspace{0.8em} Hongyuan Zhu$^2$ \\
	 $^1$Institute of High Performance Computing, A*STAR \\
	    {\tt\small \{xiuchao.sui,shaohua\}@gmail.com, \{xuxinx,liuyong,gohsm\}@ihpc.a-star.edu.sg} \\
	 $^2$Institute for Infocomm Research, A*STAR \\
	 {\tt\small\{geng\_xue,wuy,zhuh\}@i2r.a-star.edu.sg}
}

\maketitle

\begin{abstract}
Optical flow estimation aims to find the 2D motion field by identifying corresponding pixels between two images. Despite the tremendous progress of deep learning-based optical flow methods, it remains a challenge to accurately estimate large displacements with motion blur. This is mainly because the correlation volume, the basis of pixel matching, is computed as the dot product of the convolutional features of the two images. The locality of convolutional features makes the computed correlations susceptible to various noises. On large displacements with motion blur, noisy correlations could cause severe errors in the estimated flow. To overcome this challenge, we propose a new architecture ``CRoss-Attentional Flow Transformer'' (CRAFT), aiming to revitalize the correlation volume computation. In CRAFT, a Semantic Smoothing Transformer layer transforms the features of one frame, making them more global and semantically stable. In addition, the dot-product correlations are replaced with transformer Cross-Frame Attention. This layer filters out feature noises through the Query and Key projections, and computes more accurate correlations. On Sintel (Final) and KITTI (foreground) benchmarks, CRAFT has achieved new state-of-the-art performance. Moreover, to test the robustness of different models on large motions, we designed an image shifting attack that shifts input images to generate large artificial motions. Under this attack, CRAFT performs much more robustly than two representative methods, RAFT and GMA. The code of CRAFT is is available at \url{https://github.com/askerlee/craft}.
\end{abstract}

\section{Introduction}
\label{sec:intro}

\begin{figure}
\centering
  \includegraphics[scale=0.4]{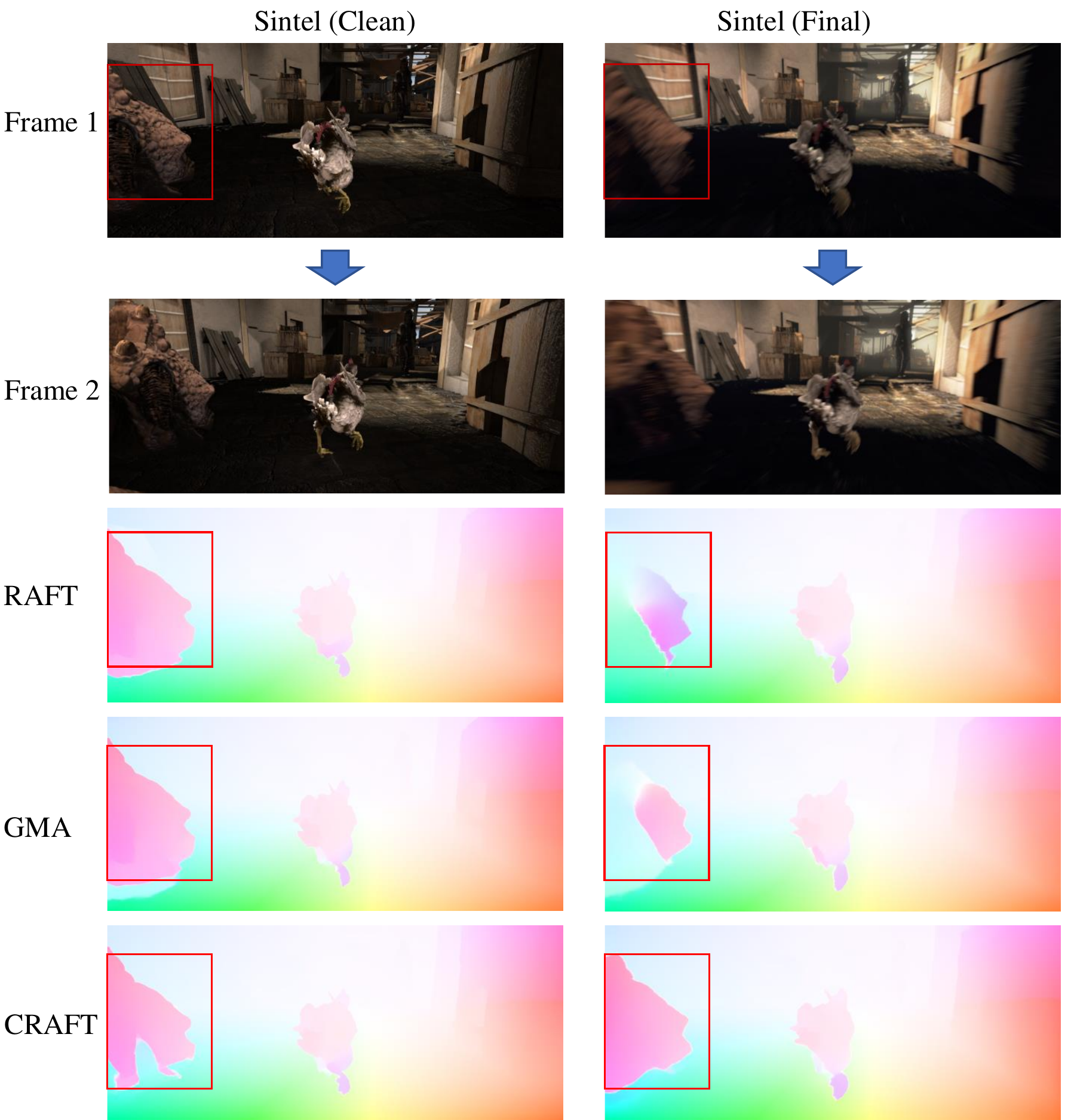}
  \captionof{figure}{The optical flow fields estimated by RAFT, GMA and CRAFT on two frames from Sintel test set, in which a dragon is chasing a chicken. On the Clean pass, all the three methods perform similarly. On the Final pass, as the area enclosed in the red rectangle has large motions ($80\sim 100$ pixels) with motion blur, RAFT and GMA only identified part of the motions. Nonetheless, CRAFT still performs well. 
  }
  \label{fig:sintel-market}
\end{figure}
~
Optical flow estimates pixel-wise 2D motions between two consecutive video frames by matching corresponding pixels. It is a fundamental computer vision task with broad applications in action recognition \cite{actor-centric,of-guided-features,integration-of}, video segmentation \cite{rigidmask,motion-grouping}, video frame interpolation \cite{super-slomo}, medical image registration \cite{flowreg}, representation learning \cite{flowe,self-video-rep}, autonomous driving \cite{kitti2}, and robot navigation \cite{flow-for-flying-robots}.

In recent years, deep learning based methods have advanced optical flow estimation tremendously \cite{flownet,flownet2,dcflow,spynet,pwc-net,maskflownet,raft,gma}. Although newest methods are very accurate on benchmark data, under certain conditions, such as large displacements with motion blur \cite{what-of-for}, flow errors could still be large. It spurs us to dig deeper to identify the root causes.

Most of these methods perform optical flow estimation based on a \emph{correlation volume} (also known as a cost volume), which stores the pairwise similarity between each pixel in Frame 1 and another in Frame 2. Given the correlation volume, subsequent modules try to match the two images, with an aim of maximizing the overall correlations between matched regions. 
The current paradigm computes the pairwise pixel similarity as the \emph{dot product} of two \emph{convolutional} feature vectors. Due to the locality and rigid weights of convolution, limited contextual information is incorporated into pixel features, and the computed correlations suffer from a high level of randomness, such that most of the high correlation values are spurious matches (Figure \ref{fig:attvis-sintel-final}). Noises in the correlations increase with noises in the input images, such as loss of texture, lighting variations and motion blur. Naturally, noisy correlations may lead to unsuccessful image matching and inaccurate output flow (Figure \ref{fig:sintel-market}). This problem becomes more prominent when there are large displacements. Reducing noisy correlations can lead to substantial improvements of flow estimation \cite{liteflownet3,sep-flow}. 

Recent years have witnessed the widespread adoption of transformers for computer vision tasks \cite{vit,detr}. An important advantage of Vision Transformers (ViTs) over convolution is that, transformer features better encode global context, by attending to pixels with dynamic weights based on their contents. For the optical flow task, useful information can propagate from clear areas to blurry areas, or from non-occluded areas to occluded areas \cite{gma}, to improve the flow estimation of the latter. A recent study \cite{vit-smooth} suggests that, ViTs are low-pass filters that do spatial smoothing of feature maps. Intuitively, after transformer self-attention, similar feature vectors take weighted sums of each other, smoothing out irregularities and high-frequency noises. 

Inspired by the feature denoising property of ViTs, we propose ``CRoss-Attentional Flow Transformer'' (CRAFT), a novel architecture for optical flow estimation. With two novel components, CRAFT revitalizes the computation of the correlation volume. First, a \emph{semantic smoothing transformer} layer fuses the features of one image, making them more global and semantically smoother. Second, a \emph{cross-frame attention} layer replaces the dot-product operator for correlation computation. It provides an additional level of feature filtering through the Query and Key projections, so that the computed correlations are more accurate.

We performed extensive evaluations of CRAFT on common optical flow benchmarks. On Sintel (Final) and KITTI (foreground) benchmarks, CRAFT has achieved new state-of-the-art (SOTA) performance. In addition, to test the robustness of different models on large motions, we designed an image shifting attack that shifts input images to generate large artificial motions. As the motion magnitude increases, CRAFT performs robustly, while two representative methods, RAFT and GMA, deteriorate severely.

\section{Related Work}
FlowNet \cite{flownet} is a pioneering work that uses deep neural networks to do end-to-end optical flow learning. It inspires a series of deep learning methods, such as FlowNet2.0 \cite{flownet2}, DCFlow \cite{dcflow}, SpyNet \cite{spynet}, PWC-Net \cite{pwc-net}, MaskFlowNet \cite{maskflownet} LiteFlowNet3 \cite{liteflownet3}, ScopeFlow \cite{scopeflow} and IRR \cite{resrefine}. Most of these methods use a \emph{correlation volume} as the basis of pixel matching.

RAFT \cite{raft} is an important development of deep learning flow methods. By using multi-scale correlation volumes and iterative flow refinement, RAFT achieves good performance, and is the precursor of a few successive works, such as GMA \cite{gma}, RAFT-Stereo \cite{raft-stereo} and CRAFT. GMA \cite{gma} is among the first works to incorporate transformer into optical flow methods. In the motion regression stage (cf.  Figure \ref{fig:craft-arch}), it uses self-attention to propagate motion features from non-occluded areas to occluded areas, and helps estimate more accurate flow of occluded areas. It complements with the improvements of CRAFT on correlation volumes.

All the aforementioned methods compute correlations using dot-product or cosine similarity of convolutional features. Within this paradigm, some works improve the efficiency of the correlation volume, such as VCN \cite{volume-corresp} and DICL \cite{disp-invariant}. Similar to our objective, Separable Flow \cite{sep-flow} aims to improve the accuracy of the correlation volume, by decomposing the 4D correlation volume into two 3D volumes, for the $u$- and $v$-directional flow regression, respectively. Separable Flow essentially imposes stronger inductive biases to obtain more accurate correlations than RAFT, as well as more accurate flow\footnote{Unfortunately, we could not compare Separable Flow with CRAFT wrt. the correlation volume accuracy, as their source code is unavailable.}. In contrast, CRAFT improves correlation computation by using contextualized frame features and reducing feature noises.

Optical flow training requires large, expensive annotated datasets. SelFlow \cite{selflow} and Autoflow \cite{autoflow} are two self-supervised methods that generate synthetic annotations. SMURF \cite{smurf} integrates a set of techniques to do self-supervised learning on unannotated video frames and has achieved promising results.

\section{The CRAFT Architecture}
\begin{figure*}[th]
\centering
  \includegraphics[scale=0.46]{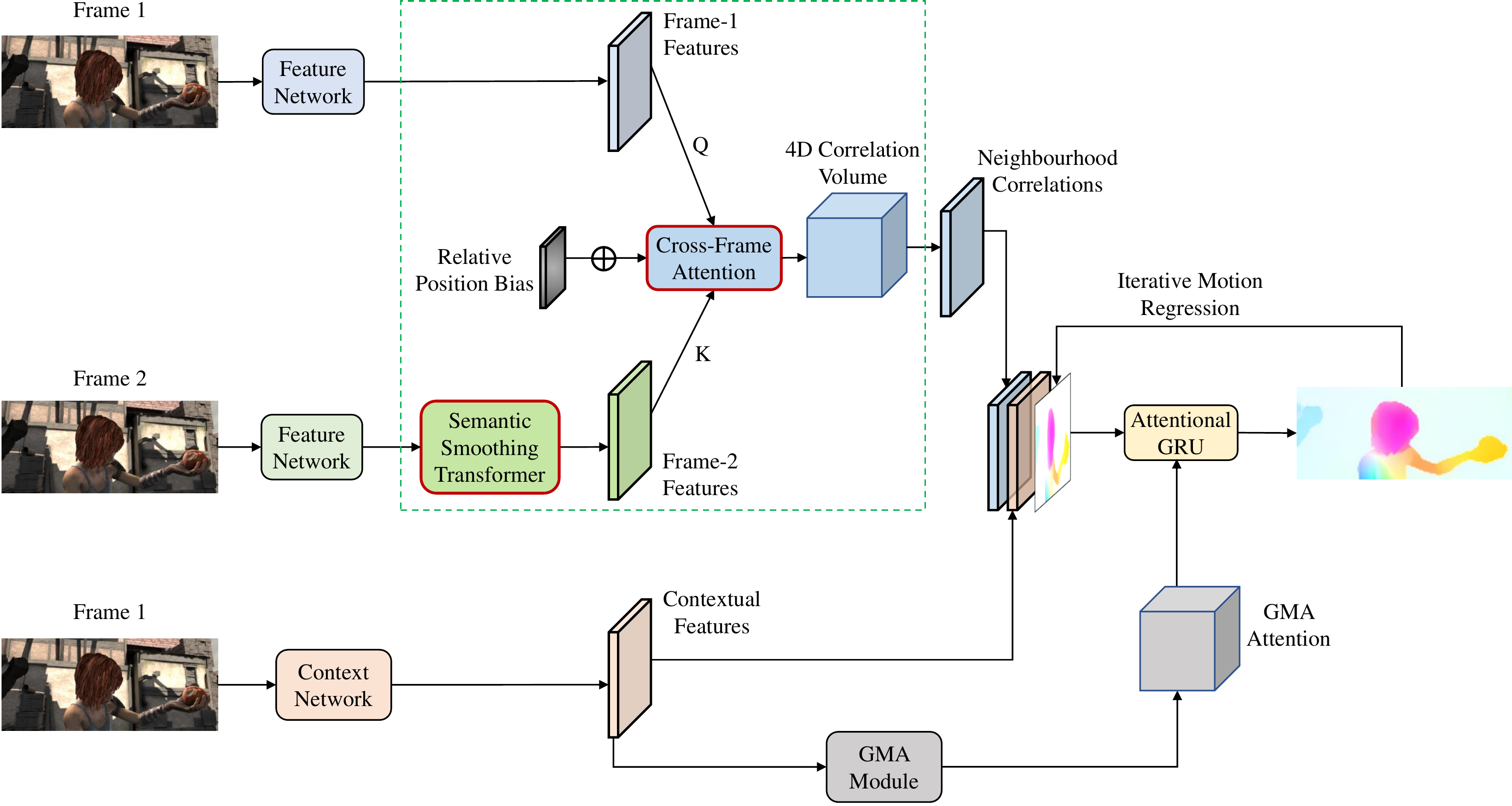}
  \captionof{figure}{CRAFT architecture. In the correlation volume computation part (the dashed green rectangle), two novel components are highlighted as boxes with red borders: the \emph{Semantic Smoothing Transformer} fuses and smooths the Frame-2 features, and the \emph{Cross-Frame Attention} layer computes the correlation volume. The GMA module at the bottom is a Global Motion Aggregation module \cite{gma}.}
  \label{fig:craft-arch}
\end{figure*}
~
Figure \ref{fig:craft-arch} presents the architecture of CRAFT. It inherits the influential flow estimation pipeline of RAFT \cite{raft}. Our main contribution is to revitalize the correlation volume computation part (the dashed green rectangle) with two novel components: the Semantic Smoothing Transformer on Frame-2 features, and a Cross-Frame Attention Layer to compute the correlation volume. These two components help suppress spurious correlations in the correlation volume, as visualized in Figure \ref{fig:attvis-sintel-final}.

\subsection{Semantic Smoothing Transformer}\label{sec:sstrans}

Given two consecutive images -- Frame 1 and Frame 2 -- as input, the first step of the flow pipeline is to extract frame features using a convolutional feature network.

To enhance the frame features with better global context, the \emph{Semantic Smoothing Transformer} (or simply \emph{SSTrans}) is used to transform the Frame-2 features. 

To better accommodate diverse features, we adopt the Expanded Attention proposed in \cite{segtran} as the SSTrans, instead of the commonly used Multi-Head Attention (MHA) \cite{transformer}. Expanded Attention is a type of Mixture-of-Experts \cite{sparse-moe} with higher capacities, and has demonstrated advantages over MHA for image segmentation tasks.

An Expanded Attention (EA) layer consists of $N$ modes (sub-transformers), computing $N$ sets of features, which are aggregated into one set using dynamic mode attention \cite{segtran}:
\begin{align}
\boldsymbol{X}_{out}^{(k)} =& \operatorname{Transformer}^{(k)}(\boldsymbol{X}), \\
\boldsymbol{B}^{(k)} =& \operatorname{Linear}^{(k)}(\boldsymbol{X}_{out}^{(k)}), \\
\text{with } &k \in \{1,\cdots,N\}, \\
\boldsymbol{G} =& \operatorname{softmax}\left(\boldsymbol{B}^{(1)}, \cdots, \boldsymbol{B}^{(N)}\right), \\
\operatorname{EA}(\boldsymbol{X}) =& \boldsymbol{G}^\top \cdot \left(\boldsymbol{X}_{out}^{(1)}, \cdots,  \boldsymbol{X}_{out}^{(N)} \right),
\end{align}
where $\boldsymbol{B}^{(k)}$ are mode attention scores, and the mode attention probabilities $G$ are softmax of all $\boldsymbol{B}^{(k)}$ along the mode dimension. The output features $\operatorname{EA}(\boldsymbol{X})$ are a linear combination of all mode features. 

To better preserve original frame features, we add a weighted skip connection with a learnable weight $w_1$:
\begin{equation}
    \operatorname{SSTrans}(\boldsymbol{X}) = w_1 \boldsymbol{X} + (1-w_1) \operatorname{EA}(\boldsymbol{X}),
\end{equation}

To impose spatial biases, we found conventional positional embeddings do not form meaningful biases, and use a \emph{relative position bias} \cite{swin,transformer-pos} instead. The bias is a matrix $B\in \mathbb{R}^{(2r+1)\times (2r+1)}$, added to the computed attention, where $r$ is the radius specifying the local range of the bias. 

Specifically, suppose the original attention matrix is reshaped to a 4-dimensional tensor $A\in \mathbb{R}^{H\times W\times H \times W}$, where $H, W$ are the height and width of the frame feature maps. For each pixel at $i,j$, where $i \in \{1,\cdots,H\}, j \in \{1,\cdots,W\}$, $A(i,j)$ is a matrix, specifying the attention weights between pixel $(i,j)$ with all the pixels in the same frame. The relative position bias $B$ is added to the neighborhood of radius $r$ of pixel $(i,j)$:
\begin{align}
    & A'(i,j,i+x,j+y) \nonumber \\ 
    = & \begin{cases}
    A(i,j,i+x,j+y) + B(x,y), & \quad \text{if } |x|\le r, |y|\le r\\
    A(i,j,i+x,j+y). & \quad \text{otherwise}
    \end{cases}
\end{align}

In our implementation, we choose the number of modes to be 4, and the radius $r$ of the relative position bias to be 7. 

\begin{figure}
\centering
  \includegraphics[scale=0.13]{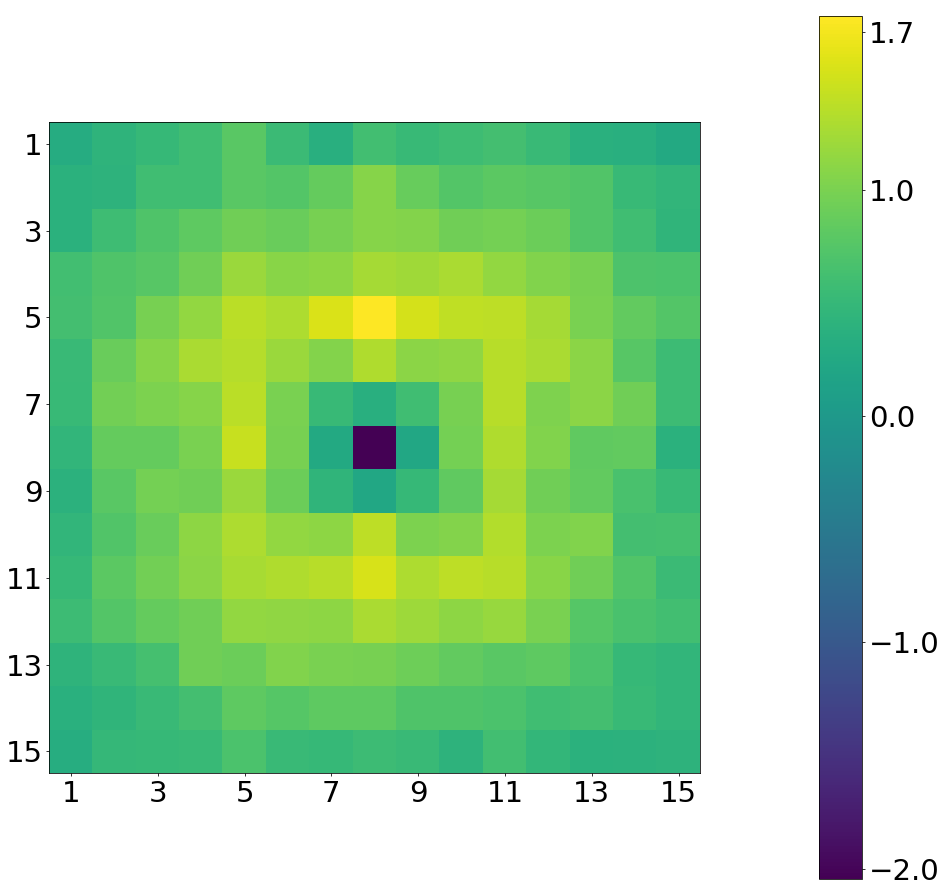}
  \captionof{figure}{Learned relative positional bias with radius $r=7$. Two interesting patterns can be observed, as detailed below.}
  \label{fig:sintel-f2bias}
\end{figure}

Figure \ref{fig:sintel-f2bias} visualizes the learned relative position bias of CRAFT trained on Sintel. Two interesting patterns are observed:
\begin{enumerate}\setlength\tabcolsep{2pt}
    \item The minimum bias value is around $-2$ located at $(0,0)$, which means that, when computing the new features of a pixel $(i,j)$, this bias term will \emph{reduce the weight of its own features} by 2. Without this term, the attention weight of pixel $(i,j)$ to itself will probably dominate the weights to other pixels, as a feature vector is most similar to itself. This term reduces the proportions of the old features of a pixel in the combined output features, effectively \emph{encouraging inflow of new information from other pixels}.
    \item The largest weights are $2\sim 3$ pixels\footnote{Here ``pixels" mean points in feature maps, which correspond to $\times 8$ pixels in the input image.} away from the center pixel, meaning that features of these surrounding pixels are most often used to supplement the features of the central pixel. 
\end{enumerate}
These two observations are confirmed in Figure \ref{fig:f2vis-sintel}, where each query draws new features from a nearby area. Setting the position bias to 0 leads to performance degradation.

It is tempting to apply transformers on the features of both frames. However, in our experiments, doing so leads to performance drop. Our hypothesis is based on the common belief that image matching heavily relies on high-frequency (HF) features that are local and structural \cite{robust-align}. Meanwhile, there are abundant HF noises that pollute informative features and hinder matching. SSTrans serves as a low-pass filter to suppress HF noises \cite{vit-smooth}, but at the same time, may reduce HF features and enhance low-frequency (LF) features. Hence, the model learns to trade off between the LF and HF components in Frame 2 for matching with Frame 1. After applying SSTrans on both frames, both frames contain less HF and more LF components. Matching them may yield many spurious correlations and hurt flow accuracy. This intuition is confirmed in Figure \ref{double-sstrans}.

\subsection{Cross-Frame Attention for Correlation Volume}
In the current paradigm, a correlation volume is the basis of cross-frame pixel matching.
After the frame features $\boldsymbol{f}_1\in \mathbb{R}^{H\times W\times D}$ and $\boldsymbol{f}_2\in \mathbb{R}^{H\times W\times D}$ are computed, the correlation volume is computed as a 4D tensor $\boldsymbol{C}\in \mathbb{R}^{H\times W\times H \times W}$ (dashed green rectangle in Figure \ref{fig:craft-arch}).

Traditionally, the correlation volume is computed as the pairwise dot-product between $\boldsymbol{f}_1$ and $\boldsymbol{f}_2$ \cite{raft}:
\begin{equation}
    C(i,j,m,n) = \frac{1}{\sqrt{D}} \boldsymbol{f}_1(i,j)^\top \cdot \boldsymbol{f}_2(m,n).
\end{equation}

Conceptually, the correlation volume is essentially Cross Attention \cite{transformer} in transformers, without feature transformation by the Query and Key projections. The query/key projections can be viewed as feature filters that separate out most informative features for correlations. In addition, to capture diverse correlations, we could use multiple query and key projections, as with Expanded Attention (EA) \cite{segtran}. Similar multi-faceted correlations are pursued in VCN \cite{volume-corresp} with multiple channels. These benefits motivate us to replace the dot-product with a simplified EA:
\begin{align}
    C_k(i,j,m,n) &= \frac{1}{\sqrt{D}} (\boldsymbol{f}_1(i,j)\boldsymbol{Q}_k)^\top \cdot \boldsymbol{K}_k\boldsymbol{f}_2(m,n), \\
    C(i,j,m,n) &= \sum_{k=1}^K \operatorname{softmax}(C_k(i,j,m,n))C_k(i,j,m,n),
\end{align}
where $\boldsymbol{Q}_k$, $\boldsymbol{K}_k$ are the $k$-th query and key projections, respectively; $C_k(i,j,m,n)$ is the correlation computed with the $k$-th mode. The $\operatorname{softmax}$ operator is taken along the $K$ modes, and aggregates the $K$ correlations. The EA here is simplified by removing the value projection and the feedfoward network. The weights of $\boldsymbol{Q}_k$ and $\boldsymbol{K}_k$ are tied, as the correlation between two frames is symmetric.

\textbf{Global correlation normalization}\hspace{1em} Sometimes extreme values may appear in the correlation volume, which may disrupt the pixel matching. To match a pixel, intuitively the relative orders of the correlations with candidate pixels are more important than absolute correlation values. In this light, we perform layer normalization \cite{layernorm} on the whole correlation volume to stabilize correlations. Empirically, this leads to slightly improved performance.


\section{Experiments}
Our experiments consist of six parts: 
\begin{enumerate}\setlength\tabcolsep{2pt}
    \item \textbf{Standard evaluation.} We evaluate different methods on Sintel \cite{sintel} and KITTI \cite{kitti}. On the two public leaderboards, CRAFT has achieved the state-of-the-art performance on both Sintel (final pass) and KITTI (foreground regions). 
    \item \textbf{Error distribution wrt. motion magnitudes.} To study the model behavior when the motion becomes larger, we calculate the flow error distribution wrt. different magnitudes of motions. CRAFT is significantly more accurate than other methods on \emph{large motions}, and performs equally well on small motions.
    \item \textbf{Ablation studies.} To analyze the impact of different components in CRAFT, i.e., the Semantic Smoothing transformer, the Cross-Frame Attention and the GMA module, we remove each of them and evaluate the ablated models on the KITTI-2015 benchmark. All these components show importance to the final performance.
    \item \textbf{Image Shifting attack.} To test the robustness of models, we manually create large motions by shifting the first frames. At very large shifts, RAFT and GMA deteriorate severely. CRAFT is significantly more robust.
    \item \textbf{Visualization of correlation volumes.} We visualize the correlations between a query point in Frame 1 and all pixels in Frame 2, to intuitively learn the differences between the correlation volumes computed by different models. CRAFT has the fewest spurious correlations compared with RAFT and GMA.
    \item \textbf{Visualization of semantic smoothing transformer attention.} To gain an intuitive idea how a pixel draws information from surrounding pixels through the SS transformer, we visualize the self-attention between a query point and all pixels in Frame 2.
\end{enumerate}

\paragraph{Training Loss} Following RAFT \cite{raft}, the loss function we adopt is a weighted multi-iteration $l_1$ loss.

\paragraph{Training Schedule} We follow the same optical flow training procedure \cite{raft,gma} of first pretraining the models on FlyingChairs (``C") \cite{flownet} for 120k iterations (batch size = 8), then on FlyingThings (``T") \cite{flyingthings} for another 120k iterations with (batch size = 6). For Sintel evaluation, we fine-tune all models on a combination of FlyingThings, Sintel (``S") \cite{sintel}, KITTI 2015 (``K") \cite{kitti} and HD1K (``H") \cite{hd1k} for 120k iterations (batch size = 6). For KITTI evaluation, we fine-tune all models on KITTI 2015 for 50k iterations (batch size = 6). Following \cite{raft,gma}, we adopt the one-cycle learning rate scheduler with the same learning rates, in which 5 percent of the iterations are used for warm-up. 

\paragraph{Evaluation Metrics}
The main evaluation metric, also used by the Sintel leaderboard\footnote{\url{http://sintel.is.tue.mpg.de/quant?metric_id=0&selected_pass=0}}, is the average end-point
error (AEPE), which is the average pixelwise flow error, measured by number of pixels. The KITTI leaderboard\footnote{\url{http://www.cvlibs.net/datasets/kitti/eval_scene_flow.php?benchmark=flow}} uses the Fl-fg (\%) and Fl-All (\%) metrics, which refer to the
percentage of outliers (pixels whose end-point error is
$> 3$ pixels or $5\%$ of the ground truth flow magnitude), averaged over foreground regions and all pixels, respectively.

\subsection{Standard Evaluation}
\setlength\tabcolsep{2pt}
\begin{table*}[t!]
\centering
\newcolumntype{C}{>{\centering\arraybackslash}X}
\begin{tabularx}{\textwidth}{@{}l l | C C C C | C C C c}
\toprule
\multirow{2}[3]{1.5cm}{Training Data} & & \multicolumn{4}{c|}{\underline{On Training Sets}} & \multicolumn{4}{c}{\underline{On Test Sets from Leaderboards}} \\
& Method & \multicolumn{2}{c}{Sintel (train)} & \multicolumn{2}{c}{KITTI-15 (train)} & \multicolumn{2}{|c}{Sintel (test)} & \multicolumn{2}{c}{KITTI-15 (test)} \\
\cmidrule(lr){3-4}
\cmidrule(lr){5-6}
\cmidrule(lr){7-8}
\cmidrule(lr){9-10}
&  & Clean & Final & AEPE & Fl-all (\%) & Clean & Final & Fl-fg (\%) & Fl-all (\%)\\
\midrule
\multirow{4}{1.5cm}{C + T / Autoflow} 
                       & RAFT\cite{raft}        & (1.43) & (2.71) & (5.04) & (17.4) & - & - & - & -\\ 
                       & RAFT-A \cite{autoflow} & (1.95) & (2.57) & (\textbf{4.23}) & - & - & - & -\\ 
                       & Perceiver-IO \cite{perceiver-io} & (1.81) & (\textbf{2.42}) & (4.98) & - & - & - & -\\ 
                       & Separable Flow\cite{sep-flow}       & (1.30) & (2.59) & (4.60) & (\textbf{15.9}) & -  & - & - & - \\
                       & GMA \cite{gma} & (1.30) & (2.74) & (4.69) & (17.1) & - & - & - & - \\
                       \cmidrule[\lightrulewidth](r{0.3em}){2-10}
                       & \textbf{CRAFT} & (\textbf{1.27}) & (2.79) & (4.88) & (17.5) & - & - & - & - \\
                       \midrule
\multirow{4}{1.5cm}{C + T + S/K + H} 
                     & RAFT\cite{raft}  & (0.76) & (1.22) & (0.63) & (1.5) & 1.61 & 2.86  & 6.87 & 5.10 \\
                     & RAFT-A \cite{autoflow}  & - & - & - & - & 2.01 & 3.14 & 5.99 & 4.78 \\
                     & RFPM \cite{RFPM} & (0.61) & (\textbf{1.05}) & (0.60) & (1.41) & 1.41 & 2.90 & - & 4.79 \\
                     & Separable Flow\cite{sep-flow} & (0.69) & (1.10) & (0.69) & (1.60) & 1.50 & 2.67 & 6.24 & \textbf{4.64} \\                     
                     & GMA \cite{gma}  & (0.62) & (1.06) & (\textbf{0.57}) & (\textbf{1.20}) & 1.39 & 2.47 & 7.03 & 5.15 \\
                     \cmidrule[\lightrulewidth](r{0.3em}){2-10}
                    & \textbf{CRAFT}  & (\textbf{0.60}) & (1.06) & (0.58) & (1.34) & 1.45 & \textbf{2.42}$^{\dagger}$ & \textbf{5.85}$^{\dagger}$ & 4.79 \\
                     \bottomrule
\end{tabularx}
\caption{\textbf{Results on Sintel and KITTI 2015 benchmarks.} We report the average end-point error (AEPE) where not otherwise stated, as well as the Fl-fg and Fl-all metrics for the KITTI dataset, which are the percentages of optical flow outliers (pixels with significant flow errors), calculated on the foreground regions and all pixels, respectively. ``C + T / Autoflow'' refers to methods that are pretrained either on the combined Chairs and Things datasets, or on the Autoflow dataset \cite{autoflow}. ``S/K + H'' refers to methods that are fine-tuned on the Sintel, KITTI and HD1K datasets. All results on Sintel (test) are generated with the ``warm-start'' strategy \cite{raft}. \\
$^\dagger$Results are ranked as the top-1 (as of November 2021) on the two public leaderboards, which include many other methods not listed here.\\
(Result) denotes a result on \emph{training sets}, listed here for reference purposes. \\
}
\label{tab:Results}
\end{table*}
Seven recent methods are compared, most of which are selected from the top-performing methods on the Sintel and KITTI leaderboards: 
\begin{itemize}
\setlength\itemsep{3pt}
    \item \textbf{RAFT} \cite{raft}: an important recent methods, and was previous SOTA before being surpassed by GMA.
    \item \textbf{RAFT-A} \cite{autoflow} uses the synthesized AutoFlow dataset (instead of ``C+T") to pretrain RAFT, followed by the standard fine-tuning steps.
    \item \textbf{Perceiver-IO} \cite{perceiver-io} is a general architecture not specifically designed for optical flow estimation. It is pretrained on Autoflow, the same as RAFT-A. The performance on the test sets is not reported in  their paper.
    \item \textbf{RFPM} \cite{RFPM} replaces the downsampling layers of RAFT to improve the flow estimation on fine details. The performance under ``C+T / Autoflow" training is not reported in their paper.
    \item \textbf{Separable Flow} \cite{sep-flow} decomposes the 4D correlation volume as two 3D volumes for the $u$ and $v$ directions. 
    \item \textbf{GMA} \cite{gma}: a recent method that enhances RAFT with a Global Motion Aggregation module to better estimate the motions of occluded pixels.
    \item \textbf{CRAFT}: with 4 modes in expanded attention layers.
\end{itemize}
    
Table \ref{tab:Results} summarizes the evaluation results of the seven methods on Sintel and KITTI. The results on the training sets (in parentheses, left side of the table) can hardly reflect how well the models generalize to new data, and are only listed for reference. The results on the test sets are evaluated on held-out data by the Sintel and KITTI servers and obtained from their leaderboards, and better reflect model performance. Although performing closely to the other methods on the training sets, CRAFT shows clear advantages on the test sets, and outperforms all other optical flow methods\footnote{As of November 2021.} on Sintel (Final) and KITTI Fl-fg (i.e., fewest foreground outliers).

We argue that these two performance metrics (AEPE on Sintel Final pass, and Fl-fg on KITTI) has important practical implications. For real world performance, the results on Sintel (Final) are more indicative than on Sintel (Clean), as the final-pass images more closely resemble real world videos, with various lighting variation, shadows and motion blur. In addition, as the foreground objects in KITTI are usually cars, pedestrians, etc., which naturally are more important than the background. Hence, smaller pixel errors in foreground regions as measured by Fl-fg, probably imply greater practical benefits than smaller errors in background.

\subsection{Error Distribution wrt. Motion Magnitudes}
\begin{table}[th]
\begin{center}
\setlength\tabcolsep{3pt}
\begin{tabular}{ll cccccc } 
\hline
 GT range & $<$~1 & [1,10] & (10,20] & (20,30] & $>$~30 & All \\
 \hline
Things-Clean \\
RAFT & {0.45} & 0.54 & 0.75 & 1.40 & 7.55 & 3.14 \\
GMA & \textbf{0.42} & \textbf{0.46} & \textbf{0.68} & {1.29} & {7.71} & 3.14 \\
CRAFT  & {0.43} & \textbf{0.46} & \textbf{0.68} & \textbf{1.26} & \textbf{6.64} & \textbf{2.77} \\ \hline
Things-Final \\
RAFT & {0.46} & 0.52 & 0.74 & 1.44 & 7.11 & 2.98 \\
GMA & \textbf{0.41} & \textbf{0.45} & {0.68} & {1.25} & {6.76} & 2.80 \\
CRAFT  & {0.42} & \textbf{0.45} & \textbf{0.65} & \textbf{1.21} & \textbf{6.11} & \textbf{2.57} \\ \hline
\end{tabular}
\caption{\textbf{AEPE on Things (validation set) in different motion ranges}. CRAFT has significantly lower AEPE on large motions.}
\label{tab:rangecomp}
\end{center}
\end{table}
To analyze the behavior of different models when facing varying magnitudes of motions, we divide the pixels into five subsets according to their groundtruth motion magnitudes, and evaluate the AEPE within each subset. As the validation/test splits of Sintel and KITTI are unavailable, the evaluation is done on the validation split of FlyingThings, Clean pass and Final pass, respectively. Three models, RAFT, GMA and CRAFT are evaluated. All the models are trained on ``C+T". 

Table \ref{tab:rangecomp} presents the AEPE on different magnitudes of motions. When the motion is $<$~20 pixels, CRAFT performs on par with GMA. On large motions that are $>30$ pixels, CRAFT makes 10$\sim$15\% less AEPE than RAFT and GMA.

\subsection{Ablation Studies}
\begin{table}[th]
\begin{center}
\setlength\tabcolsep{4pt}
\begin{tabular}{l cc } 
\hline
 KITTI-15 (test) & Fl-fg (\%) & Fl-all (\%) \\
 \hline
CRAFT & 5.85 & 4.79 \\ \hline
-SS trans & 6.41 & 5.06 \\
-CFA  & 6.15 & 4.90 \\ 
-GMA & 6.21 & 4.93 \\ \hline
\end{tabular}
\vspace{5pt}
\caption{\textbf{Ablated models on KITTI-2015 (test) leaderboard}.}
\label{tab:ablation}
\end{center}
\end{table}
CRAFT has three important components: the Semantic Smoothing transformer  (``SS trans"), the Cross-Frame Attention (``CFA"), and the GMA module. To study their individual contributions, in each turn we remove one of them, train the ablated models with the standard schedule, and evaluate on the KITTI-2015 leaderboard.

Table \ref{tab:ablation} shows that all the three components make important contributions to the overall performance. 

\subsection{Image Shifting Attack}
\begin{figure}
\centering
  \includegraphics[scale=0.34]{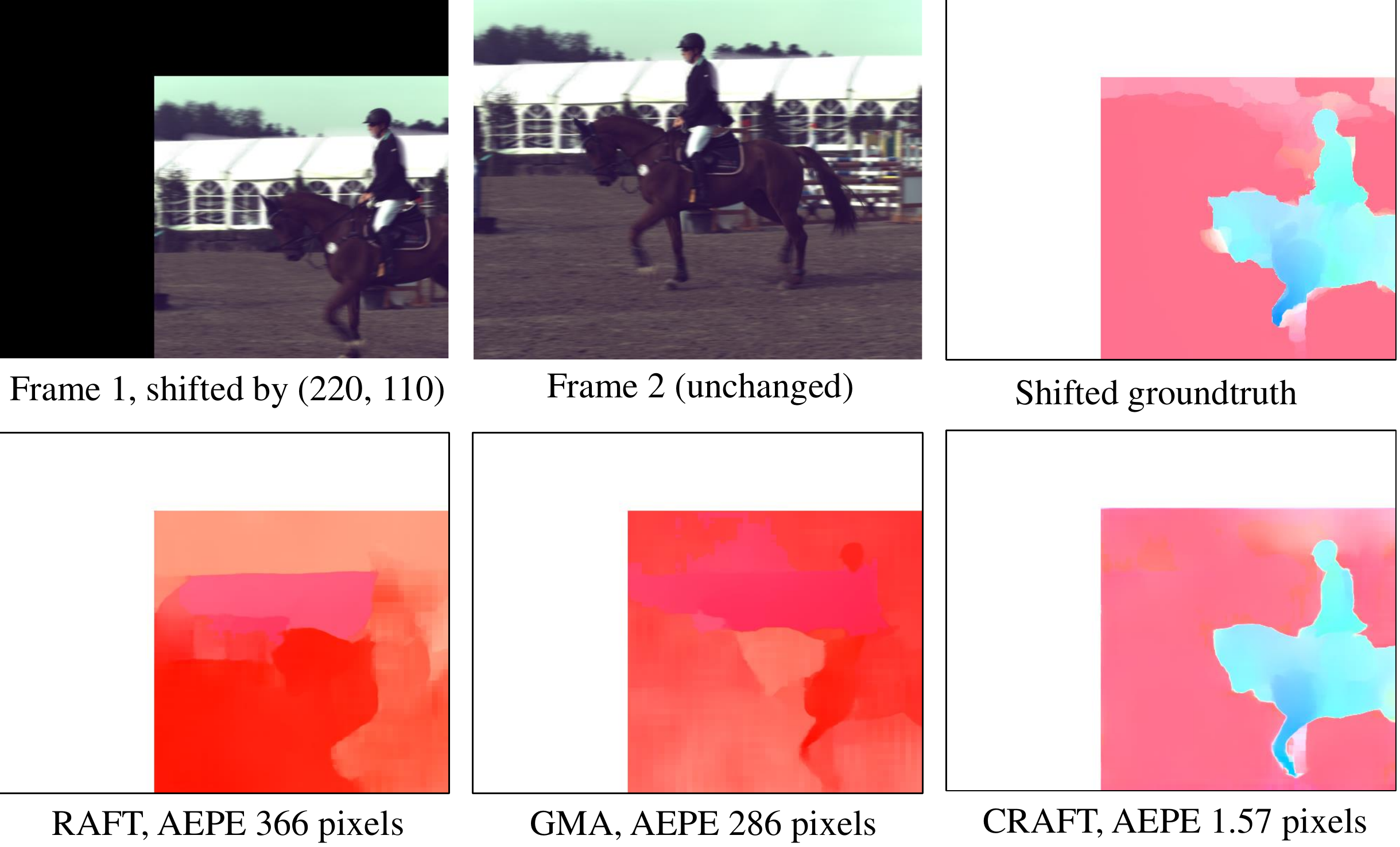}
  \captionof{figure}{Flows fields estimated by RAFT, GMA and CRAFT on two frames from the Slow Flow dataset. $(\Delta u, \Delta v)=(220, 110)$ pixels. RAFT and GMA failed with huge AEPE. CRAFT still yielded accurate estimation.}
  \label{fig:slowflow-animal}
\end{figure}
\begin{figure*}[t]
\centering
  \includegraphics[scale=0.54]{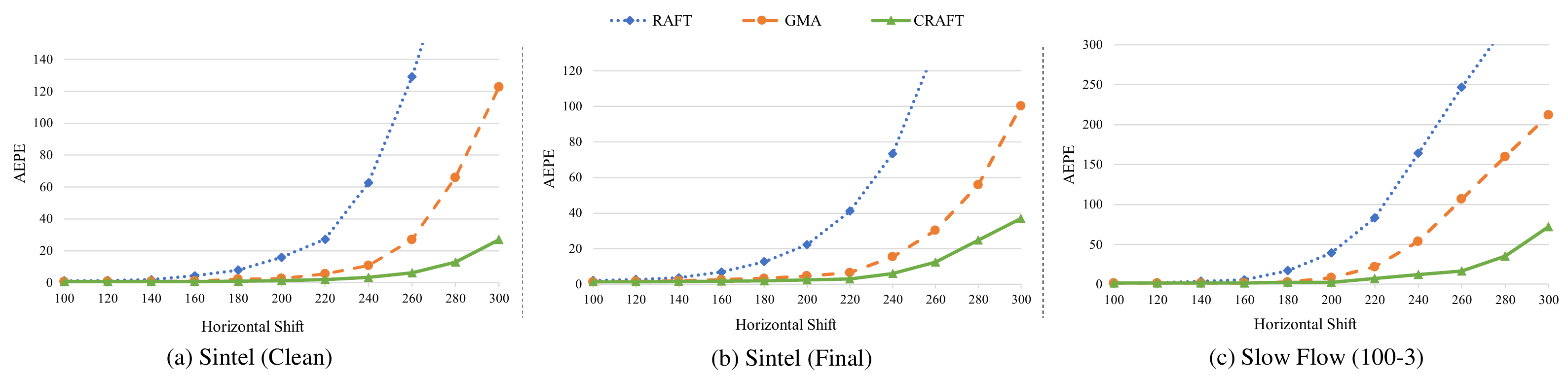}
\caption{The AEPE of RAFT, GMA and CRAFT change differently with the magnitude of image shifts. (a)-(c) are on Sintel (Clean), Sintel (Final) and Slow Flow, respectively. The horizontal shift $\Delta u$ change from 100 to 300, and the vertical shift $\Delta v \doteq \frac{1}{2}\Delta u$. When $\Delta u$ goes beyond 160, RAFT and GMA quickly deteriorate, and CRAFT performs much more robustly.}
\label{fig:shift-3datasets}
\end{figure*}

Typically, most pixels in standard benchmark images are with small motions, and large motions only appear in local areas. As a result, when the model makes big errors on large local motions, as these errors are local, they may be easily corrected by considering the contextual small motions, so that the final flow may still be accurate. Thereby, the fragility on large motions is hidden under small AEPE. 

To fully reveal the model robustness on large motions, we design an image shifting attack, i.e., create large motions by shifting one image along the $u,v$ plane. Local corrections would hardly work on such image pairs, as all the pixels will have large displacements.

Specifically, we shift the first frame $I_1$ by $(\Delta u, \Delta v)$ towards the bottom right, getting a new image $\text{shift}_{u,v}(I_1)$. The new image is truncated at the original image boundary. 

Suppose a model $M$ estimates the flow $F_0$ accurately on the original image pairs: $F_0=M(I_1, I_2) \approx F_{gt}$, where $F_{gt}$ is the groundtruth flow. We test $M$ on the shifted pairs and get new flow: $F_1=M(\text{shift}_{u,v}(I_1), I_2)$. 
Then we unshift $F_1$ and get $F_2$. If the model is robust against the shift, it can be proven that the following equation should hold:
\begin{equation}
    F_2\approx \text{shift}_{u,v}(F_0) - (\Delta u, \Delta v) \approx \text{shift}_{u,v}(F_{gt}) - (\Delta u, \Delta v).
\end{equation}

Figure \ref{fig:slowflow-animal} presents an example of the shifting attack. The two frames are from Slow Flow \cite{slowflow}, a dataset with motion blur (flow magnitude=100, blur duration=3). After downsampling the original images from (1280, 720) to (640, 360), the first image is shifted by $(220, 110)$. RAFT and GMA completely fail to estimate the flow, with huge AEPE. In contrast, CRAFT still yields accurate estimation.

Figure \ref{fig:shift-3datasets} presents the quantitative evaluations of RAFT, GMA and CRAFT under the shifting attack. The models are trained with ``C+T+S+K+H", and evaluated on the training split of Sintel (Clean) and Sintel (Final), as well as on Slow Flow (flow magnitude=100, blur duration=3), under varying $(\Delta u, \Delta v)$. In our experiments, the horizontal shift $\Delta u \in [100, 300]$, and the vertical shift $\Delta v \doteq \frac{1}{2}\Delta u$. When $\Delta u \le 160$, all models perform well with AEPE $<8$. When $\Delta u$ goes beyond 160, RAFT and GMA quickly deteriorate; in contrast, CRAFT performs much more robustly with significantly smaller AEPE. Possibly due to motion blur, the AEPE of RAFT and GMA on Slow Flow is $80\sim 100$ pixels larger than on Sintel, while the AEPE of CRAFT on Slow Flow is only 35 pixels larger, showing its robustness against motion blur.

\subsection{Visualization of Correlation Volumes}

The main reason that CRAFT performs more robustly is probably that the computed correlation volumes contains much fewer spurious correlations, thanks to the SS transformer and the cross-frame attention layer. 

To gain an intuitive understanding of the differences between the correlation volumes computed by different models, we visualize the correlations between a query point in Frame 1 and all pixels in Frame 2. The query point is marked as a small red square in Frame 1 (projected to the small green square in Frame 2). It moves to the small red square in Frame 2. The dashed green rectangle is a $256\times 256$-pixel square centered at the query point, truncated at the image boundary. It encloses the field of view (FoV) of the model at the first iteration of flow estimation. Only correlations within the FoV are shown. 

\begin{figure}[ht]
\centering
  \includegraphics[scale=0.35]{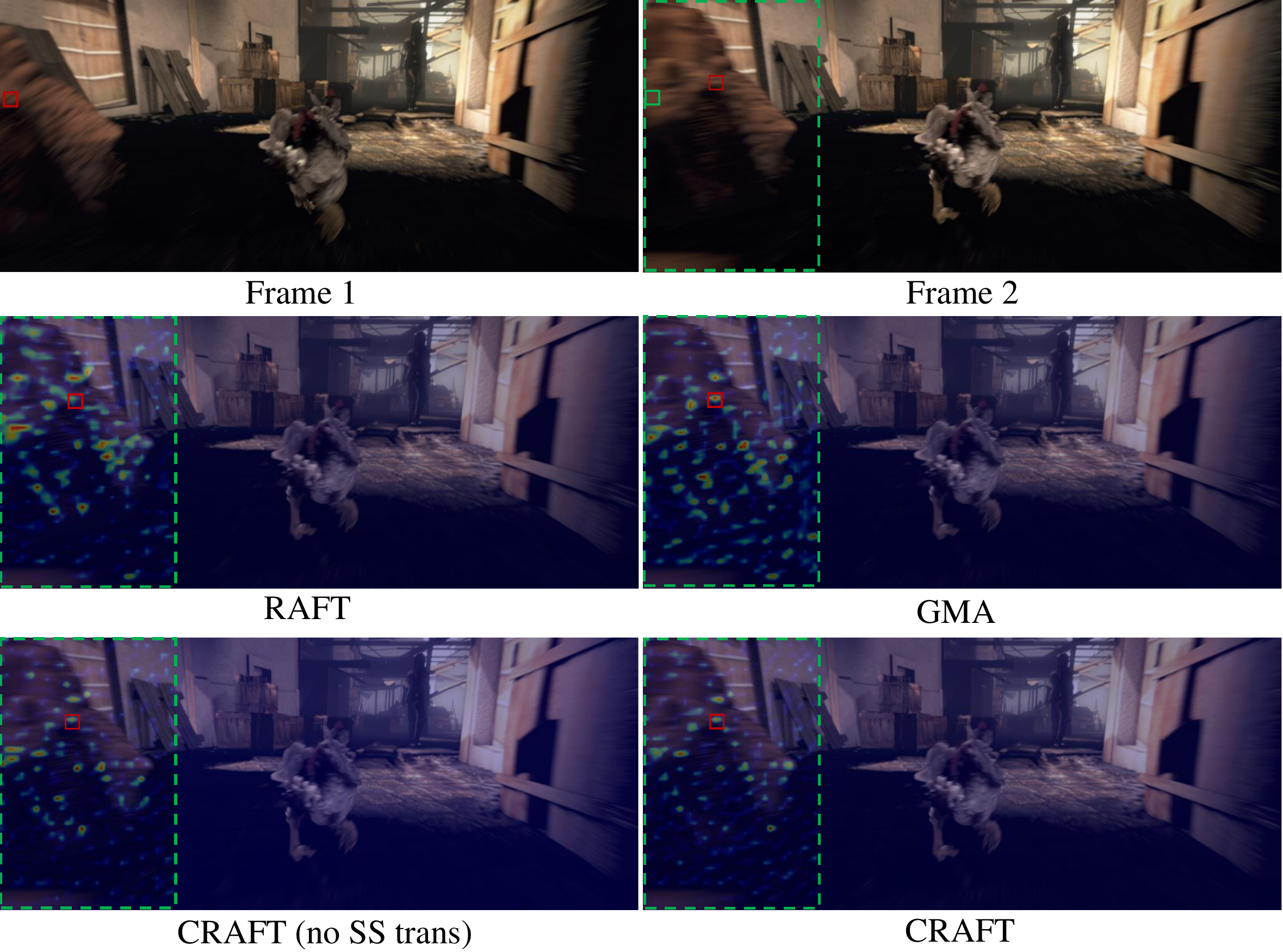}
  \caption{Heatmaps of the correlations between Frame 2 and a query point in Frame 1 (the small red square), on Sintel test set (Final pass). The small green square in Frame 2 indicates the original position of the query in Frame 1. As the images are blurry with coarser details, RAFT and GMA make many noisy correlations. In contrast, CRAFT has significantly fewer noisy correlations.}
  \label{fig:attvis-sintel-final}
\end{figure}

Figure \ref{fig:attvis-sintel-final} visualizes the correlation volumes on two frames from Sintel (Final), which is rendered with shadows and motion blur. 
Bright blobs in the heatmaps are high correlations, and those not at the groundtruth location (red square) are spurious and may be targets for mismatch. The correlation volumes\footnote{All matrices have been normalized into $[0,1]$ to make sure the pattern differences are not caused by range discrepancy.} computed by RAFT and GMA contain many more spurious correlations than CRAFT. 
If removing the SS transformer (the cross-frame attention layer remains), CRAFT yields more noisy correlations, but they are still fewer than RAFT and GMA, suggesting that the cross-frame attention layer also helps denoising.

\begin{figure}[ht]
\centering
  \includegraphics[scale=0.4]{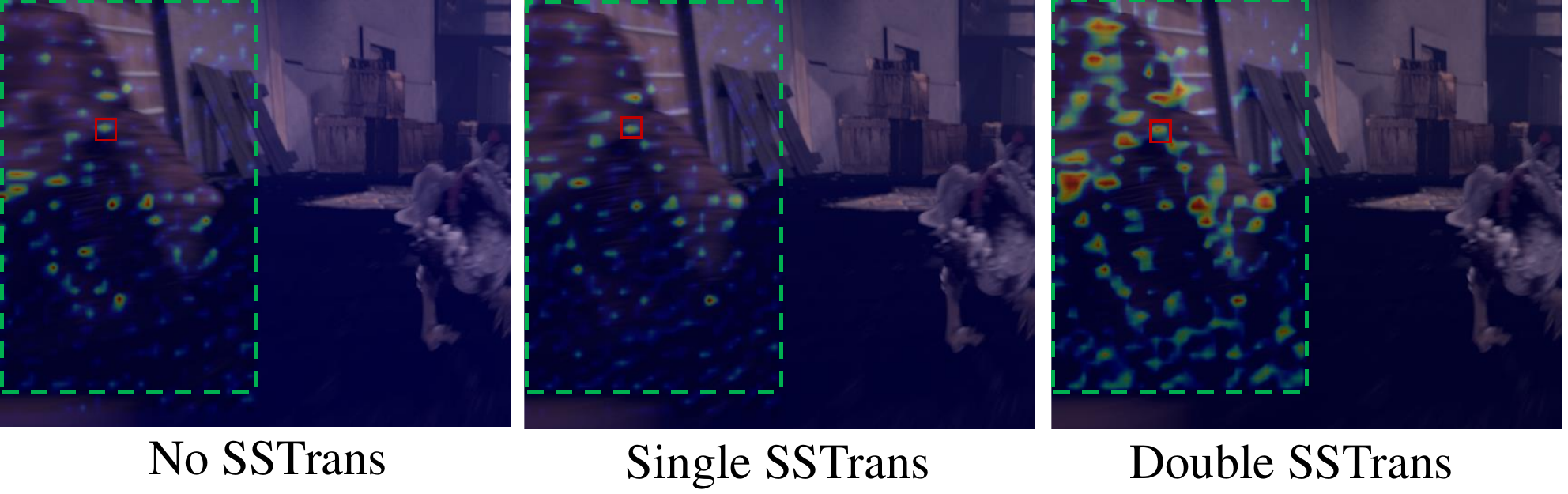}
  \caption{The correlations between Frame 2 and a query point in Frame 1, on Sintel test set (Final pass). Images are cropped. The standard CRAFT setting (``Single SSTrans") has fewest noisy correlations. ``Double SSTrans" yields many more noisy correlations.}
  \label{double-sstrans}
\end{figure}

In addition, as stated in Section \ref{sec:sstrans}, we tested to apply the SS transformer to both Frame 1 and Frame 2 (referred to as ``Double SSTrans"), and observed degraded performance. To shed light on why this happens, Figure \ref{double-sstrans} visualizes the computed correlations with Double SSTrans. Compared with the standard ``Single SSTrans", many more spurious correlations are observed. This may explain the degradation of the flow accuracy.

~
~
~ 
\begin{figure}[ht!]
\centering
  \includegraphics[scale=0.31]{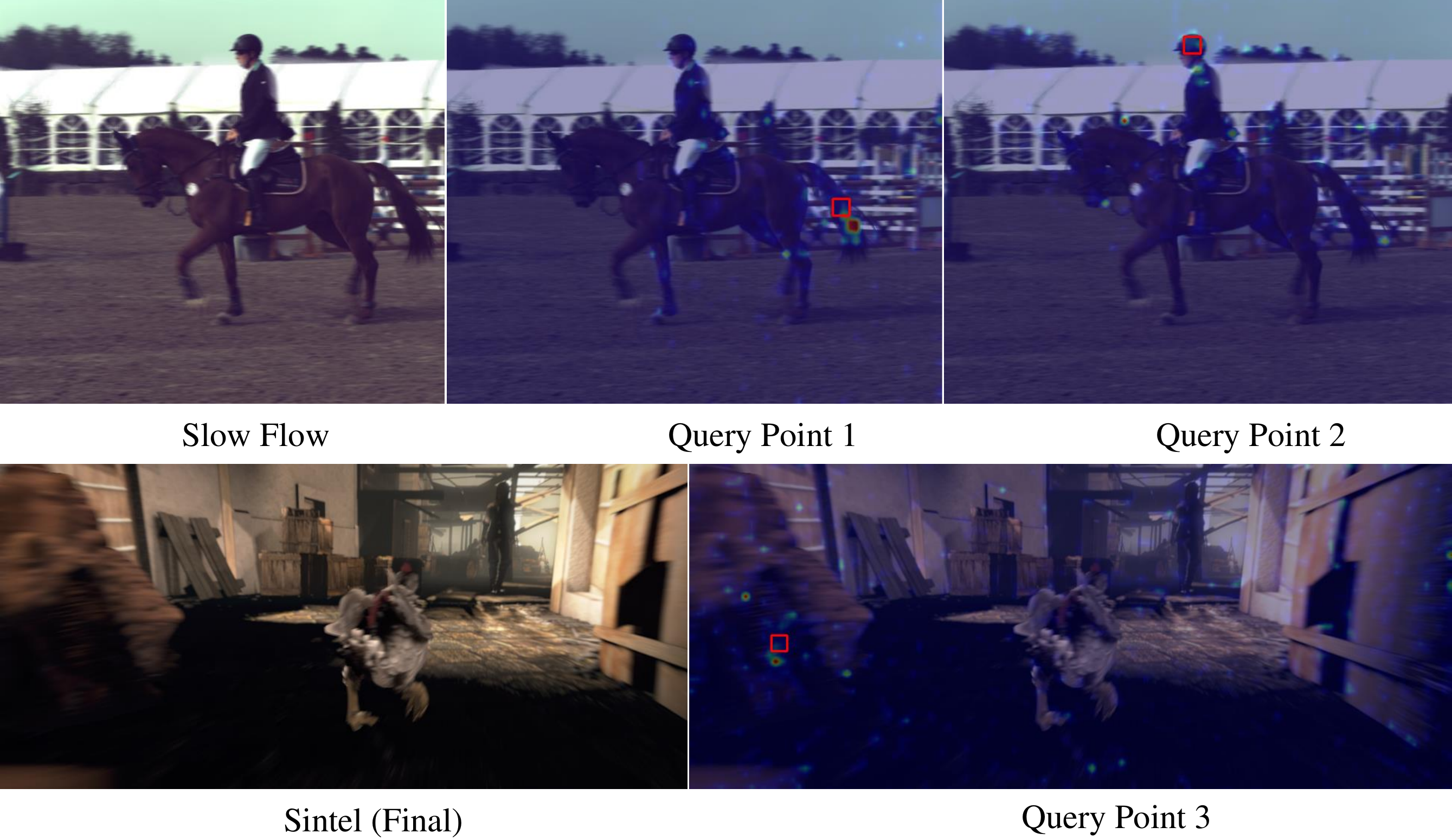}
  \caption{Heatmaps of the SS transformer self-attention, between a query point (a red rectangle) and all pixels in the same image. The most intense areas are where the query points pay the highest attention and draw features to enrich themselves.}
  \label{fig:f2vis-sintel}
\end{figure}

\subsection{Visualization of the Self-Attention of Semantic Smoothing Transformer}

Figure \ref{fig:f2vis-sintel} visualizes the SS transformer self-attention weights on three queries in Frame 2. For each query (the small red square), its attention weights with all pixels in the \emph{same image} are displayed as a heatmap. The highest attention areas are somewhere around the query points (at different relative directions). We guess that these areas may provide texture or contextual information absent at the queries.


\section{Conclusions}
We present a novel optical flow estimation method \emph{Cross-Attentional Flow Transformer} (CRAFT). It revitalizes the computation of correlation volumes with two novel components: Semantic-Smoothing Transformer and Cross-Frame Attention. They help compute more accurate correlation volumes by spatially smoothing feature semantics and filtering out feature noises. CRAFT has achieved new state-of-the-art performance on a few metrics, and is especially robust on large displacements with motion blur. 

\section*{Acknowledgements}
This research is supported by A*STAR under its Career Development Fund (Grant Nos. C210812035 and C210112016), and its Human-Robot Collaborative AI for Advanced Manufacturing and Engineering programme (Grant No. A18A2b0046).

{\small
\bibliographystyle{ieee_fullname}
\bibliography{craft}
}

\clearpage
\appendix
\section{Model Size and FLOPs}

Table \ref{flops} presents the number of parameters and FLOPs of RAFT, GMA and CRAFT. The ``Ratio" columns take RAFT as the base. FLOPs are measured while inference on Sintel images (1024x436 pixels). 

Similar to the SS Transformer and the Cross-Frame Attention, the GMA module in CRAFT is implemented with Expanded Attention \cite{segtran}, which contains multiple modes. In Table \ref{flops}, two CRAFT models with different numbers of GMA modes $m$ are presented. These two models have very similar overall performance (Table \ref{shift-results}).
When $m=2$, the model size is only 7\% and 19\% larger than GMA and RAFT, respectively, and thus the performance gain is not to be explained away as merely having more parameters.
\begin{table}[H]
\begin{centering}
\begin{tabular}{m{1.95cm}|L|c|L|c}
\hline
             & Params (M) & Ratio   & FLOPs (G) & Ratio \tabularnewline \hline 
RAFT         & 5.3        & 1       & 369       & 1     \tabularnewline 
GMA          & 5.9        & 1.11    & 494       & 1.34  \tabularnewline 
CRAFT$_{m=2}$       & 6.3        & 1.19    & 613       & 1.66 
\tabularnewline
CRAFT$_{m=4}$       & 6.3        & 1.19    & 794       & 2.15
\tabularnewline \hline 
\end{tabular}
\caption{Number of parameters / FLOPs (on Sintel images). $m$ of CRAFT denotes the number of modes in the GMA module.}\label{flops}
\par\end{centering}
\end{table}

As shown in Table \ref{comp-flops}, the GMA module dominates the total overhead of the three transformer modules. This drastic difference is because the GMA module is applied in every iteration of the iterative motion refinement \cite{raft}, while the other two are only applied once. In this regard, for the FLOPs computation above, we fixed SS Trans and CFA to have 4 modes, and only varied the number of modes of the GMA module.
\begin{table}[H]
\begin{centering}
\begin{tabular}{m{1.8cm}|c|c|c|c}
\hline
             & SS Trans & CFA   & GMA & Remaining  \tabularnewline \hline 
FLOPs (G) & 66    & 6.6       & 317 & 405
\tabularnewline \hline 
\end{tabular}
\caption{FLOPs (on Sintel images) of different components in CRAFT$_{m=4}$. SS Trans, CFA and GMA all have 4 modes.}\label{comp-flops}
\par\end{centering}
\end{table}

\section{Image Shifting as Augmentation}

To explore how manually shifting some training images impacts the model performance, we take it as an extra augmentation, namely ``ShiftAug", for the training of optical flow models. 

\subsection{Mild ShiftAug}\label{mild-shiftaug}
In the mild ShiftAug training, 10\% of the training batches are shifted by $(\Delta u, \Delta v)$ sampled from two Laplacian distributions with scales 16 and 10 (the mean value of a Laplacian distribution is the scale\footnote{https://en.wikipedia.org/wiki/Laplace\_distribution}), respectively.  

We trained two GMA models and two CRAFT$_{m=2}$ models, with and without ShiftAug, respectively. They are denoted as GMA, GMA-shift, CRAFT and CRAFT-shift.

\subsubsection{Leaderboard Evaluation}

\begin{table*}[ht!]
\centering
\begin{tabular}{c|LLL|LLL|L|L|L}
\hline 
\multirow{2}{*}{Settings} & \multicolumn{6}{c|}{Sintel} & \multicolumn{3}{c}{KITTI}\tabularnewline
\cline{2-10} \cline{3-10} \cline{4-10} \cline{5-10} \cline{6-10} \cline{7-10} \cline{8-10} \cline{9-10} \cline{10-10} 
 & \hspace{5em} All & Clean s10-40 & \hspace{5em} s40+ & \hspace{5em} All & Final s10-40 & \hspace{5em} s40+ & Fl-bg (\%) & Fl-fg (\%) & Fl-all (\%)\tabularnewline
\hline 
CRAFT$_{m=4}$ & 1.45 & 0.97 & 8.30 & 2.43 & 1.74 & \textbf{13.27} & 4.58 & \textbf{5.85} & 4.79 \tabularnewline
CRAFT$_{m=2}$ & 1.44 & 0.99 & 8.13 & \textbf{2.42} & \textbf{1.62} & 13.66 & \textbf{4.30} & 6.87 & 4.73 \tabularnewline
CRAFT-shift$_{m=2}$ & \textbf{1.40} & \textbf{0.96} & \textbf{7.84} & 2.51 & 1.73 & 14.02 & 4.35 & 6.35 & \textbf{4.68} \tabularnewline
\hline 
\end{tabular}
\caption{\textbf{Additional results on Sintel and KITTI 2015 leaderboards.} We report the average end-point error (AEPE) for Sintel, and the Fl-bg, Fl-fg and Fl-all metrics for KITTI, which are the percentages of optical flow outliers (pixels with significant flow errors), calculated on the foreground regions and all pixels, respectively. To show the performance on large motions, we present the AEPE on s10-40 and s40+, i.e., pixels whose velocities are within [10, 40] pixels, and $> 40$ pixels, respectively.}
\label{shift-results}
\end{table*}

Table \ref{shift-results} presents the performance scores of the two CRAFT$_{m=2}$ models (with or without ShiftAug), along with the standard CRAFT$_{m=4}$ (without ShiftAug), on Sintel and KITTI leaderboards. 

Without ShiftAug, when $m$ reduces from 4 to 2, the performance on large motions (``s40+" for Sintel, and ``Fl-fg" for KITTI) degrades slightly. As expected, ShiftAug recovers the model performance on large motions on Sintel (Clean) and KITTI. However, it is surprising to see that with ShiftAug, the performance on large motions on Sintel (Final) degrades slightly.

Due to the restricted frequency of submissions to the leaderboards, we were unable to evaluate more model settings before the camera ready deadline, such as CRAFT$_{m=4}$ and GMA trained with ShiftAug.

\subsubsection{Performance under Image Shifting Attack}
\begin{figure*}[ht]
\centering
  \includegraphics[scale=0.6]{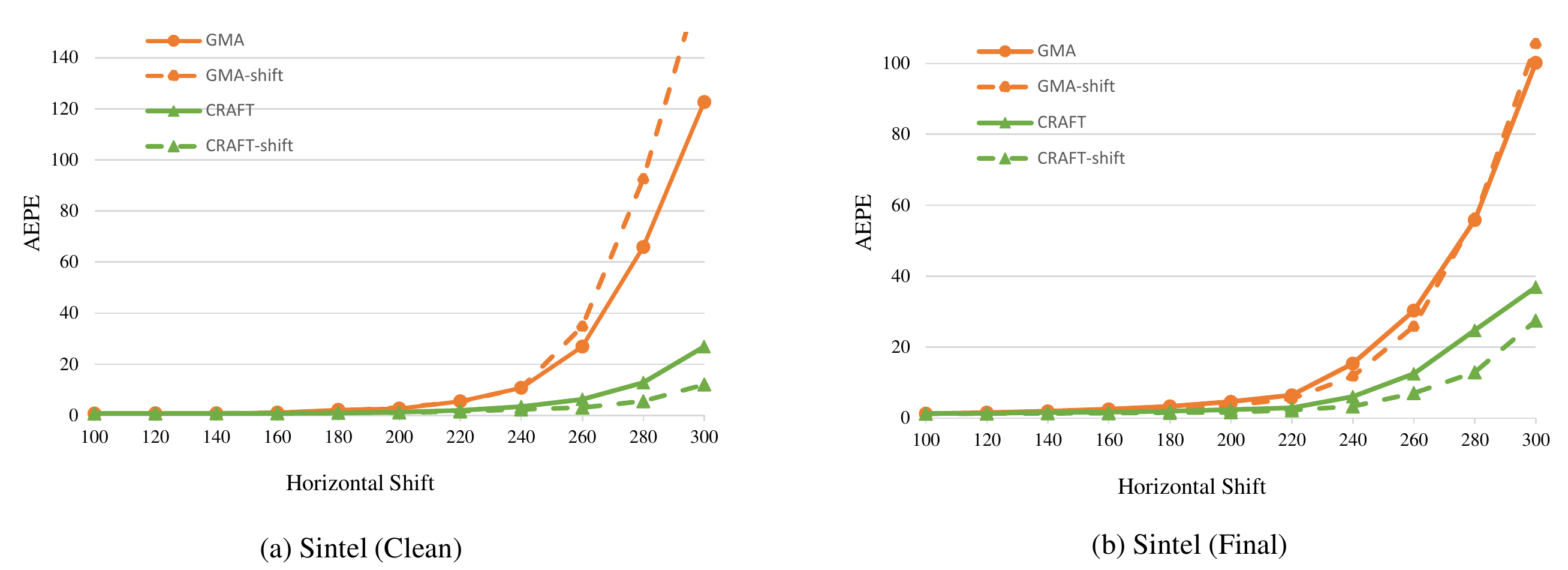}
\caption{The AEPE of RAFT, GMA and CRAFT change differently with the magnitude of image shifts. (a)-(c) are on Sintel (Clean), Sintel (Final) and Slow Flow, respectively. The horizontal shift $\Delta u$ change from 100 to 300, and the vertical shift $\Delta v \doteq \frac{1}{2}\Delta u$. When $\Delta u$ goes beyond 160, RAFT and GMA quickly deteriorate, and CRAFT performs much more robustly.}
\label{fig:shift-models}
\end{figure*}

We evaluated the 4 models under the image shifting attack, to study whether ShiftAug makes models more robust against it. Figure \ref{fig:shift-models} presents the performance of of the four models, evaluated on the training split of Sintel (Clean) and Sintel (Final), under varying $(\Delta u, \Delta v)$. The horizontal shift $\Delta u \in [100, 300]$, and the vertical shift $\Delta v \doteq \frac{1}{2}\Delta u$. 

It can be seen that, under the image shifting attack, GMA-shift and CRAFT-shift follow similar performance curves as GMA and CRAFT, respectively. On both Sintel (Clean) and Sintel (Final), CRAFT-shift yields significant smaller AEPE (20-50\%) on very large shifts ($\ge 200$ pixels). However, on Sintel (Final), GMA-shift yields almost identical AEPE on very large shifts as GMA. Bewilderingly, on Sintel (Clean), the AEPE of GMA-shift becomes significantly higher on very large shifts than GMA. ShiftAug helps both GMA and CRAFT reduce AEPE on medium-to-large shifts (120-180 pixels). Without ShiftAug, the maximum AEPE in this range is 3.4 pixels. 

Based on the observations above, we conclude that mild ShiftAug does not help make GMA more robust against very large image shifts. On the other hand, mild ShiftAug already helps make CRAFT significantly more robust against very large image shifts.

\subsection{Aggressive ShiftAug against Image Shifting}\label{aggress-shiftaug}
\begin{table}[th]
\begin{center}
\setlength\tabcolsep{4pt}
\begin{tabular}{lc cccccc} 
\hline
 $\Delta x$ (px) & 0      & 240    & 280    & 320    & 360    & 400 \\
 \hline
GMA    & 1.19   & 120.2    & 270.2   & 362.7   & 427.9   & 478.1 \\
CRAFT  & \textbf{1.11}   & 10.8   & 51.8   & 138.4 & 236.5 & 336.9 \\ \hline
GMA-shift       & 1.21 & 1.83   & 3.11   & 9.33   & 47.5   & 124.3 \\
CRAFT-shift  & 1.13   & \textbf{1.59}   & \textbf{1.74}   & \textbf{2.15} & \textbf{6.73} & \textbf{32.0} \\ \hline
\end{tabular}\vspace{-6pt}
\caption{\textbf{AEPE on Chairs under image shifting attack}.}
\end{center}
\end{table}

In this section, we aim to test the effects of more aggressive ShiftAug, i.e., with a probability of 10\%, shift frame 1 by $(\Delta x, \Delta y)$, where $\Delta x$ is uniformly drawn from $[-320, 320]$, and $\Delta y$ is uniformly drawn from $[-160, 160]$. We fine-tuned CRAFT and GMA on Things models (pretrained on C+T) with such random shifting. Then we evaluated the two models on Chairs\footnote{We intentionally chose Chairs for evaluation, which has a slight domain gap with Things (used for training), to see how the robustness generalizes.} with varying degree of shifting. GMA-shift becomes more robust against image shifting attack, but CRAFT-shift still outperforms GMA-shift with a large margin on larger shifts. 

\subsection{Innate and Acquired Robustness}
Based on the observations in Section \ref{mild-shiftaug} and \ref{aggress-shiftaug}, we hypothesize that there are two types of model robustness -- \emph{innate} and \textit{acquired} robustness. The latter is learned through augmentation, but stronger innate robustness of CRAFT makes it robust to variations beyond the training data. Hence, CRAFT is likely more robust against other unseen image variations as well, e.g., rotations, lighting variations and motion blur.

\section{Iterative Motion Refinement on Shifted Slow-Flow Images}
\begin{figure*}[ht]
\centering
  \includegraphics[scale=0.4]{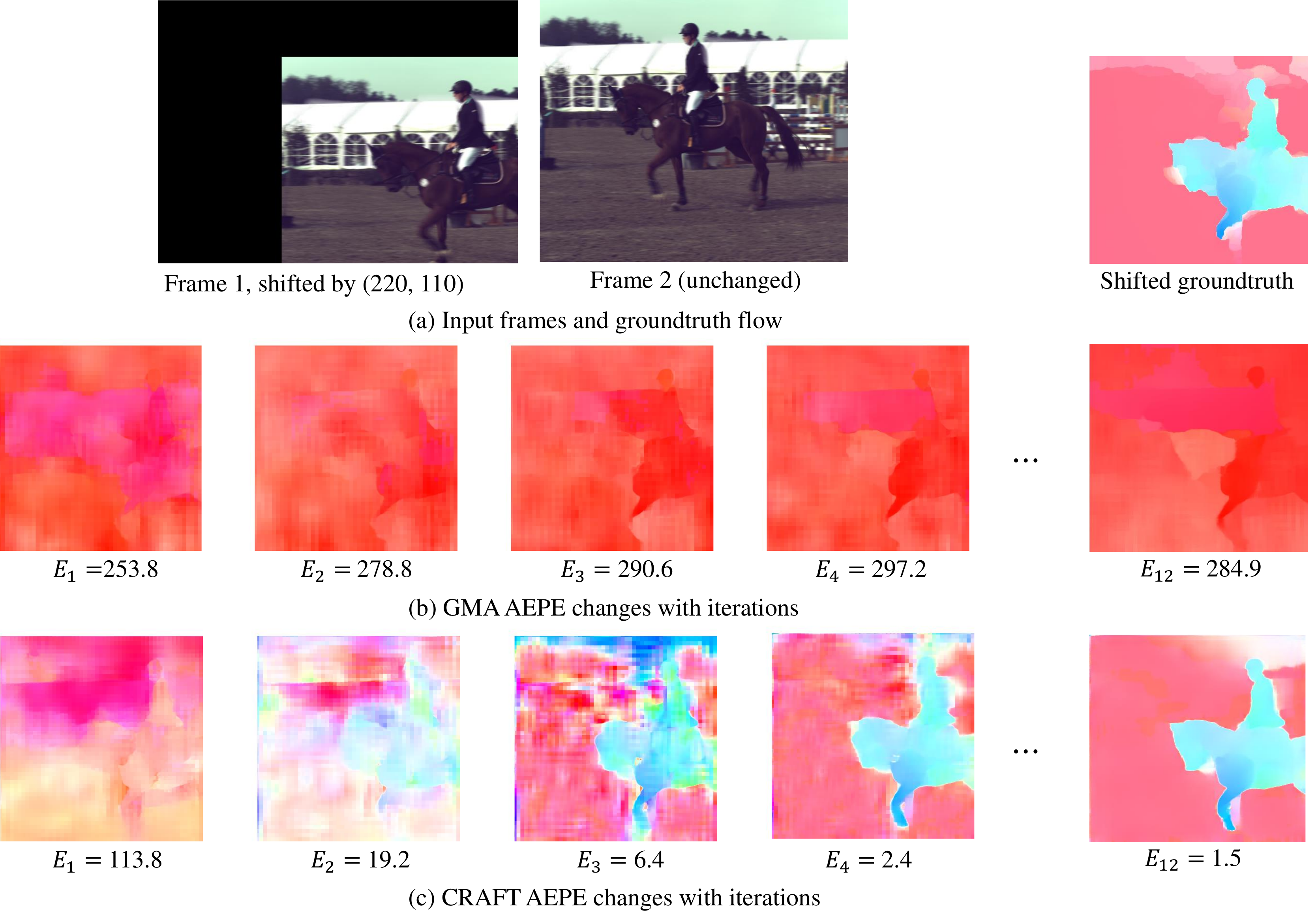}
  \caption{The iterative refinement of optical flow on the shifted Slow Flow image pair (Figure 4 in the main text), by GMA and CRAFT, respectively. As in the first iteration, GMA makes excessively huge errors, it is unable to recover with more iterations. CRAFT recovers from smaller initial errors and yields an accurate flow field eventually.}
  \label{fig:slowflow-err-by-iter}
\end{figure*}

Figure \ref{fig:slowflow-err-by-iter} presents the flow fields estimated at different iterations by GMA and CRAFT (both without shift augmentation), respectively, on the same Slow Flow \cite{slowflow} image pair as in Figure 4 of the main text. It partially explains why the AEPE (average end-point error) of GMA is huge when the image shift is large (Figure 5 in the main text). 

The flow field is estimated through iterative refinement of $N=12$ iterations in both GMA and CRAFT. At the $i+1$-th iteration, it uses the flow estimated at the $i$-th iteration as initialization, and attempts to estimate a more accurate flow field. This is effective when the flow errors are confined in very small areas, in which cases the model can correct the errors by considering  the estimated motions of the surrounding pixels (which are largely accurate). However, if large errors appear in broader areas, the model may fail to recover from the errors with more iterations. Therefore, a relatively small AEPE at the first iteration is crucial for achieving a small AEPE after all iterations. In Figure \ref{fig:slowflow-err-by-iter}, the flow estimated by GMA at the first iteration has huge errors (AEPE = 253.8), and thus even with more iterations, GMA is unable to recover. In contrast, the flow estimated by CRAFT at the first iteration has a much smaller AEPE = 113.8, and CRAFT quickly corrects the errors.

\section{Visualization of Correlation Volumes on Slow Flow}
\begin{figure*}
\centering
  \includegraphics[scale=0.5]{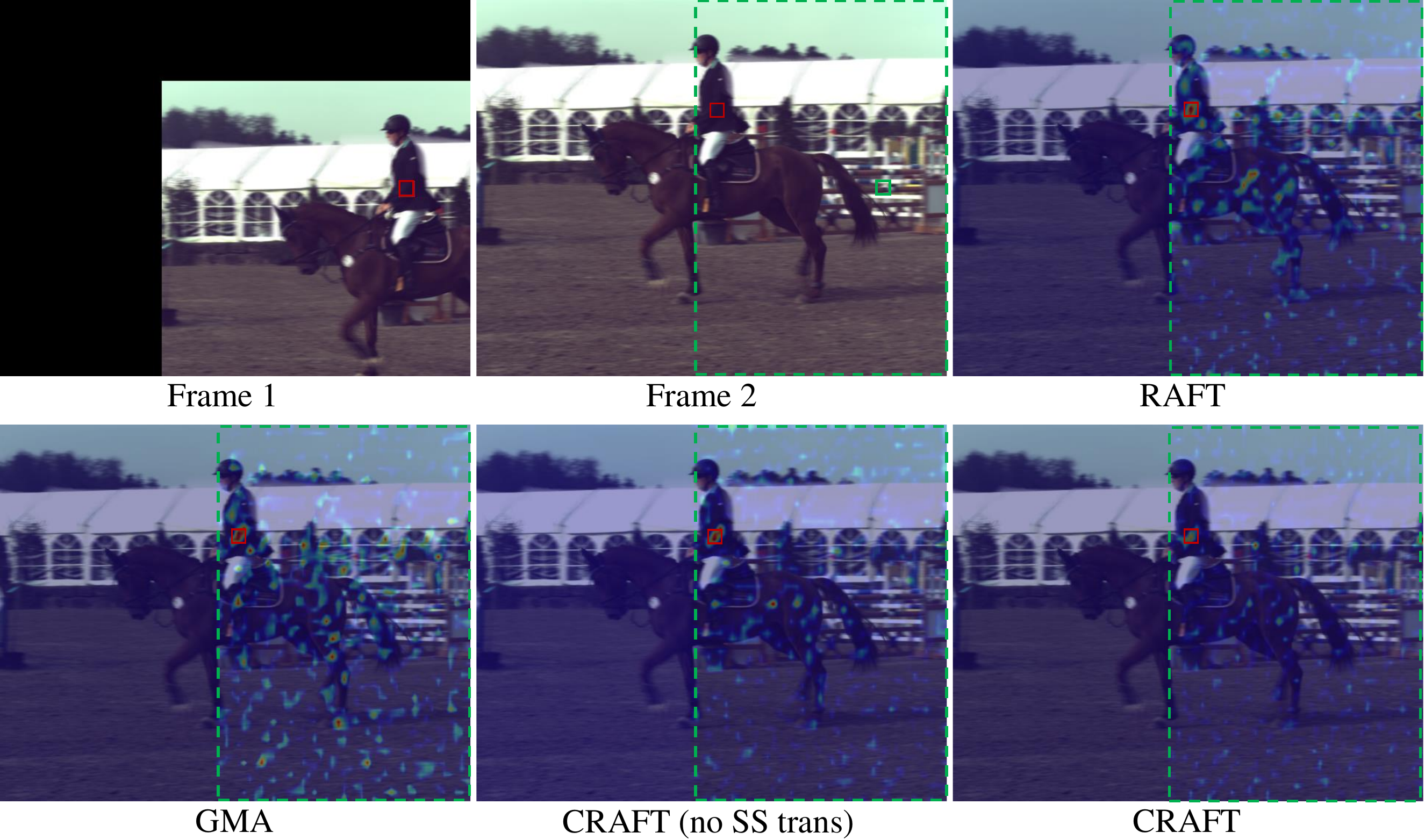}
  \caption{Heatmaps of the correlation matrices between Frame 2 and a query point on the rider's body in a shifted Frame 1, on the Slow Flow dataset. At the presence of motion blur, CRAFT has significantly fewer noisy correlations than RAFT and GMA, showing its robustness. Removing SS transformer results in more noisy correlations.}
  \label{fig:cfa-shift-slowflow-human}
\end{figure*}

\begin{figure*}
\centering
  \includegraphics[scale=0.5]{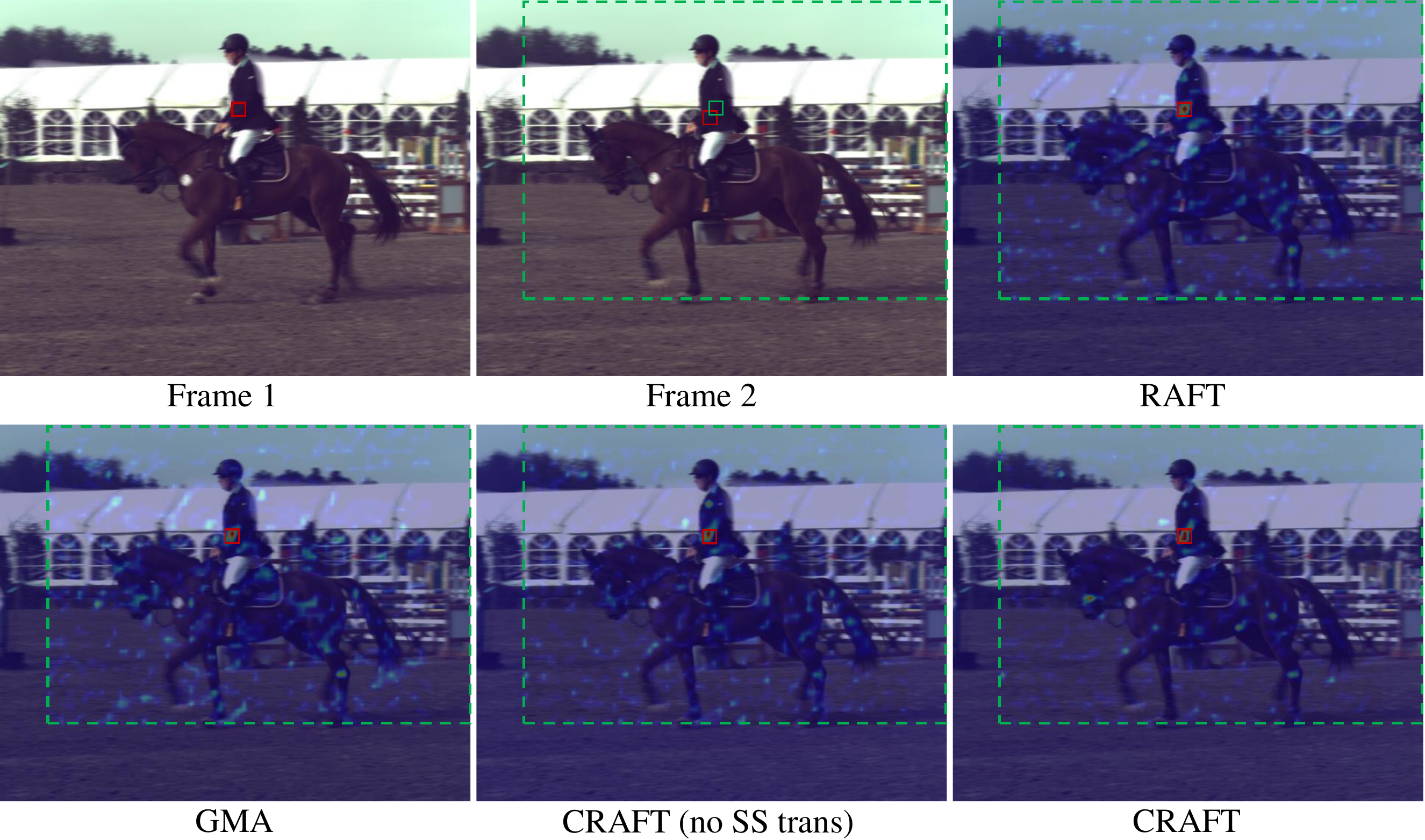}
  \caption{Heatmaps of the correlation matrices between Frame 2 and a query point on the rider's body in Frame 1, on Slow Flow.}
  \label{fig:cfa-slowflow-human}
\end{figure*}

\begin{figure*}
\centering
  \includegraphics[scale=0.5]{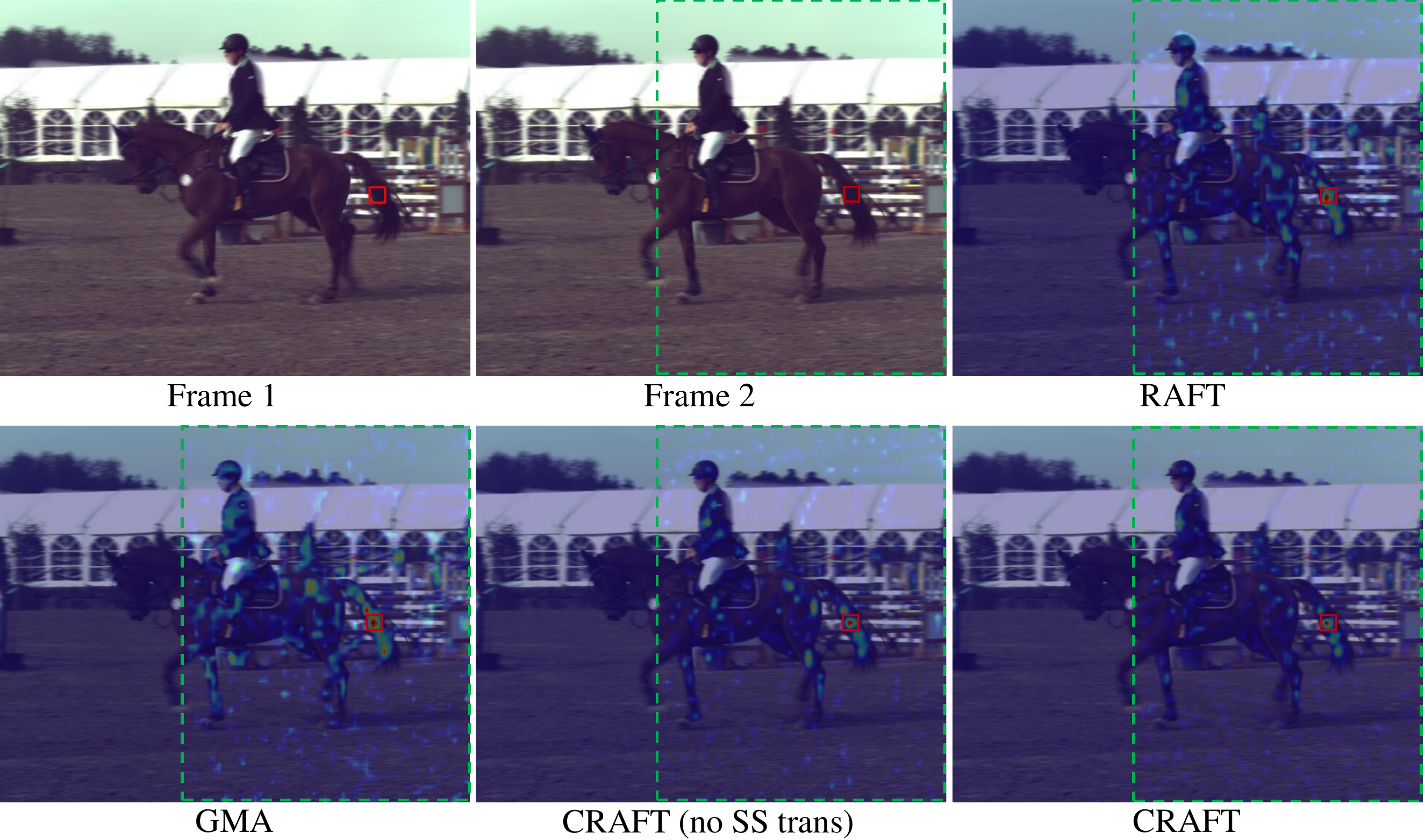}
  \caption{Heatmaps of the correlation matrices between Frame 2 and a query point on the horse's tail in Frame 1, on the Slow Flow dataset.}
  \label{fig:cfa-slowflow-horsetail}
\end{figure*}

Figures \ref{fig:cfa-shift-slowflow-human}-\ref{fig:cfa-slowflow-horsetail} present more visualizations of the correlation volume.
Figure \ref{fig:cfa-shift-slowflow-human} visualizes the correlation volume of RAFT, GMA and CRAFT on the shifted image pair from Slow Flow. This is the same example as in Figure \ref{fig:slowflow-err-by-iter}, and Figure 4 in the main text. It can be seen that, GMA has the most spurious high correlations, and CRAFT has the least. 

Figures \ref{fig:cfa-slowflow-human} and \ref{fig:cfa-slowflow-horsetail} visualize the correlation volumes with two different query points in Frame 1, on the rider's body, and on the horse's tail, respectively, on the original image pair from Slow Flow. Similar distributions of spurious high correlations are observed. Among the four models, CRAFT (with SS trans) always has the least spurious high correlations, showing that it is able to greatly suppress spurious correlations and compute a more accurate correlation volume, which may explain its robustness demonstrated in Figure \ref{fig:slowflow-err-by-iter}.

\section{Screenshots of Sintel and KITTI Leaderboards}
\begin{figure*}[ht]
    \centering 
    \includegraphics[scale=0.48]{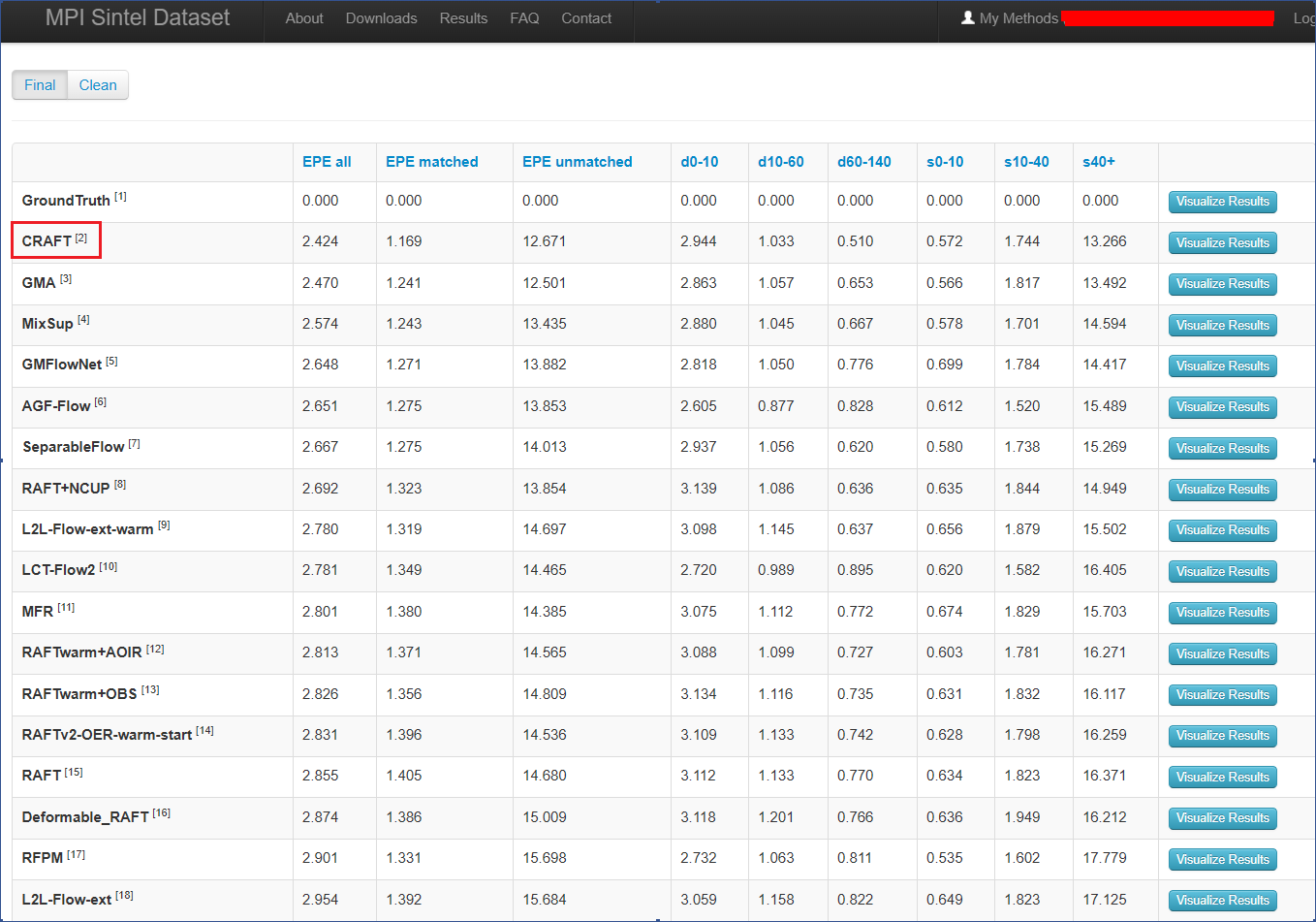}
    \caption{Screenshot of Sintel (Final) leaderboard, taken near the CVPR'2022 submission deadline.}
    \label{fig:screenshot-sintel-final}
\end{figure*}

\begin{figure*}[ht]
    \centering
    \includegraphics[scale=0.48]{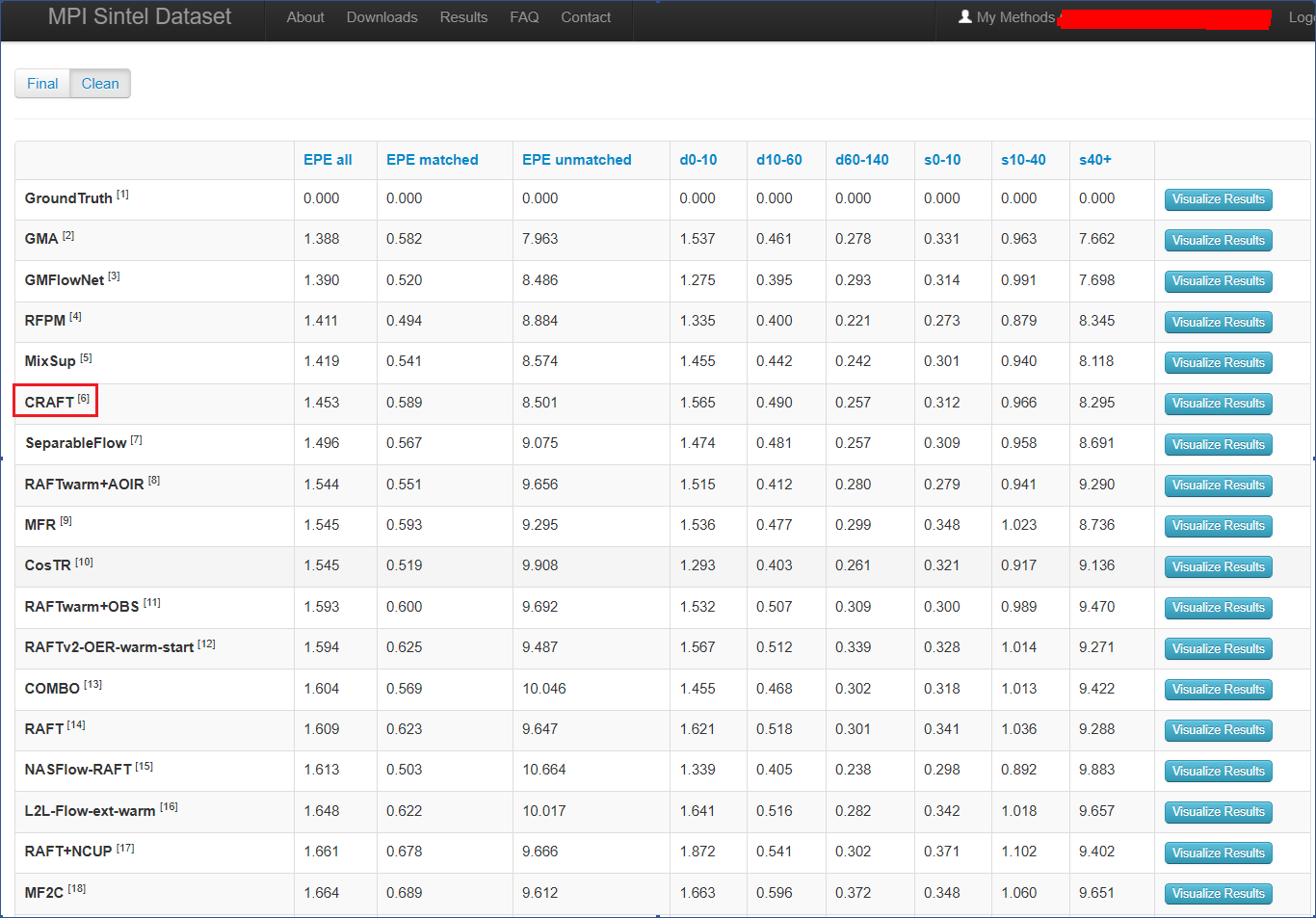}
    \caption{Screenshot of Sintel (Clean) leaderboard, taken near the CVPR'2022 submission deadline.}
    \label{fig:screenshot-sintel-clean}
\end{figure*}

\begin{figure*}[ht]
    \centering 
    \includegraphics[scale=0.9]{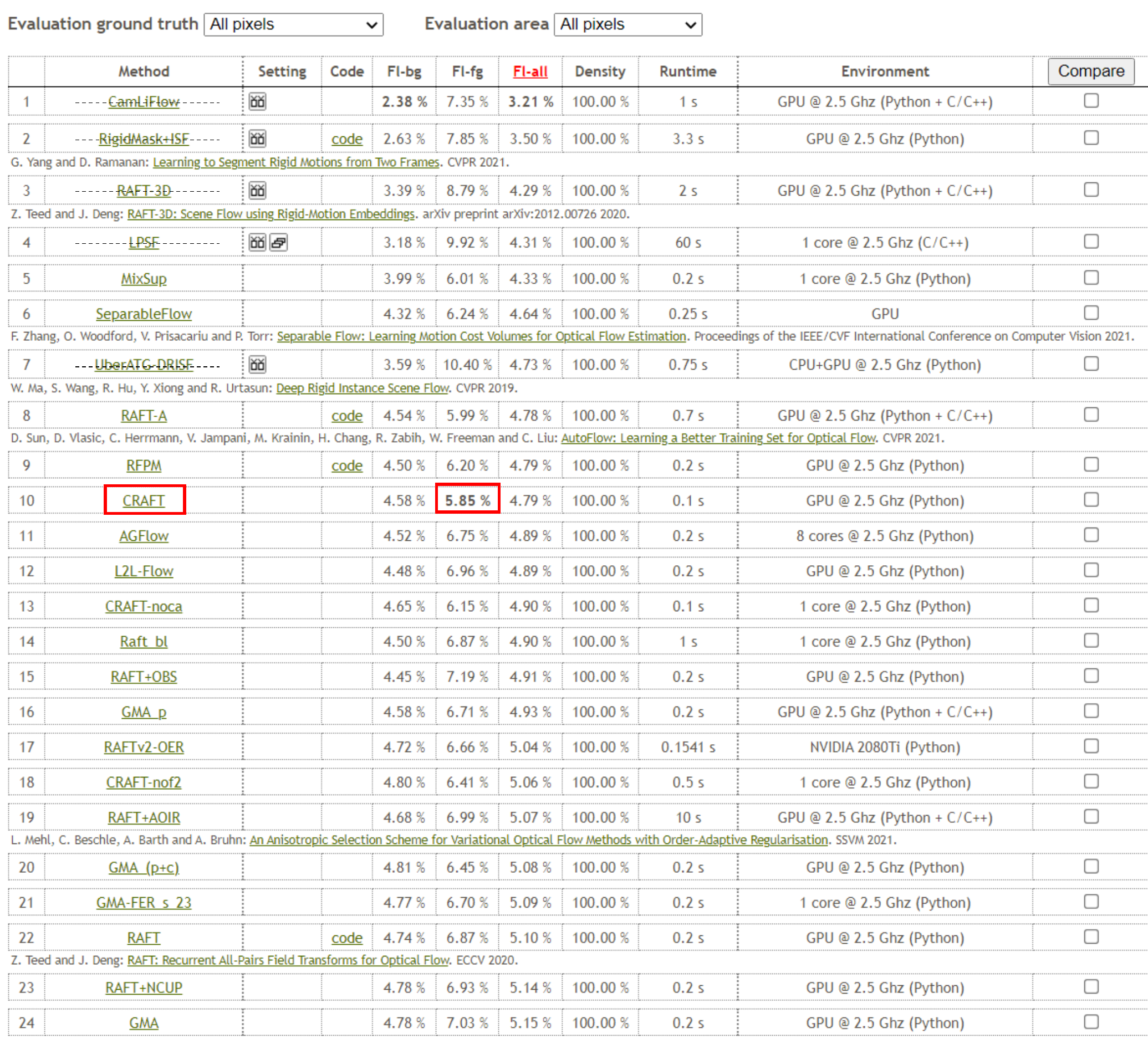}
    \caption{Screenshot of KITT-2015 leaderboard, taken near the CVPR'2022 submission deadline. Five scene flow methods are marked with \st{strikethrough text}, as they are not comparable to optical flow methods. There remain the top 19 optical flow methods. ``CRAFT-noca" and ``CRAFT-nof2" are the ablated models of removing the Cross-Frame Attention, and the Semantic Smoothing Transformer from CRAFT, respectively. }
    \label{fig:screenshot-kitti}
\end{figure*}

Figures \ref{fig:screenshot-sintel-final}-\ref{fig:screenshot-kitti} are the screenshots of the Sintel (Final), Sintel (Clean) and the KITTI-2015 optical flow leaderboards, which were taken near the CVPR'2022 submission deadline. 

CRAFT ranked the 1st and 5th places on Sintel (Final pass), Sintel (Clean pass), respectively. As Sintel (Final pass) images contain more light variations, shadows, motion blurs, etc. that are common in real world, we argue that the performance on Sintel (Final) better reflects the performance of a model on real-world images. Evidence has been presented in the Sintel paper (Figure 5, \cite{sintel}) that Sintel (Final) has similar image and motion statistics as other real-world datasets, including Lookalikes \cite{sintel} and Middlebury \cite{middlebury}.

On the KITTI flow-2015 leaderboard, a few methods among the top are scene flow methods (marked with \st{strikethrough text}) that take two stereo pairs of images as input (cf. two monocular images of optical flow), and thus are not comparable with optical flow methods. Among the top optical flow methods, CRAFT ranks 5th. In particular, it achieves the highest accuracy on foreground regions, measured as the smallest Fl-fg (percentage of flow outliers\footnote{Pixels whose end-point error is $> 3$ pixels or $5\%$ of the ground truth flow magnitude.} in the foreground regions). On Fl-all (percentage of flow outliers in both foreground and background regions), MixSup ranks as the top-1 optical flow method, but its training and implementation details are missing for further analysis and comparison with CRAFT. Separable Flow\cite{sep-flow} and RFPM\cite{RFPM} have significantly worse performance on foreground regions. RAFT-A uses Autoflow\cite{autoflow} as the pretraining data, and thus not directly comparable with CRAFT. Autoflow explore a new way to synthesise training data, which is orthogonal to our method or other recent architectures. It is worth noting that, the foreground objects in KITTI are usually cars, pedestrians, etc., which naturally are more important than the background. Thus, smaller Fl-fg are probably more important for practical applications than smaller Fl-bg or Fl-all.
\clearpage


\end{document}